\RequirePackage{amsmath}
\PassOptionsToPackage{table}{xcolor}
\documentclass[]{fairmeta}

\usepackage{amsmath,amsfonts,bm}

\def\eqref#1{equation~\ref{#1}}

\def\1{\bm{1}}

\DeclareMathAlphabet{\mathsfit}{\encodingdefault}{\sfdefault}{m}{sl}
\SetMathAlphabet{\mathsfit}{bold}{\encodingdefault}{\sfdefault}{bx}{n}

\usepackage{pifont}
\usepackage{makecell} 
\usepackage{siunitx}  
\usepackage{algorithm}
\usepackage{algpseudocode}
\usepackage{amssymb}
\usepackage{enumitem}

\usepackage{listings}
\tcbuselibrary{listings, skins, breakable}
\lstdefinelanguage{json}{
    basicstyle=\ttfamily\footnotesize,
    numbers=none,
    numberstyle=\tiny,
    stepnumber=1,
    numbersep=5pt,
    showstringspaces=false,
    breaklines=true,
    frame=single,
    backgroundcolor=\color{gray!5},
    string=[s]{"}{"},
    morestring=[b]',
    literate=
     *{0}{{{\color{blue}0}}}{1}
      {1}{{{\color{blue}1}}}{1}
      {2}{{{\color{blue}2}}}{1}
      {3}{{{\color{blue}3}}}{1}
      {4}{{{\color{blue}4}}}{1}
      {5}{{{\color{blue}5}}}{1}
      {6}{{{\color{blue}6}}}{1}
      {7}{{{\color{blue}7}}}{1}
      {8}{{{\color{blue}8}}}{1}
      {9}{{{\color{blue}9}}}{1}
}

\usepackage[normalem]{ulem}

\definecolor{lightgray}{gray}{0.4}
\definecolor{darkblue}{RGB}{94,110,186}
\definecolor{darkGreen}{RGB}{92, 148, 110}
\definecolor{darkgreen}{RGB}{0,120,0}
\definecolor{lightorange}{RGB}{255,160,0}
\definecolor{darkred}{RGB}{139,0,0}
\definecolor{yellow}{RGB}{255,255,230}
\definecolor{OliveGreen}{rgb}{0.36, 0.71, 0.39}

\newcommand{\methodname}{EgoP\textsuperscript{2}C}

\title{EgoP\textsuperscript{2}C: Internalizing Plan-Perceive-Compose for Spatial Reasoning in Egocentric Video}

\author[1,2,*]{Chaoyu Li}
\author[2]{Arka Sadhu}
\author[2]{Yuliang Li}
\author[2]{Hyo Jin Kim}
\author[2]{Yuning Chai}
\author[1]{Pooyan Fazli}

\affiliation[1]{Arizona State University}
\affiliation[2]{Meta}

\contribution[*]{Work done during an internship at Meta.}

\abstract{%
  \begingroup
  \renewenvironment{abstract}{}{}%
  \begin{abstract}

Answering complex spatial reasoning questions from egocentric video requires determining what evidence is needed, acquiring targeted observations, and combining them through multiple reasoning steps. Existing vision-language models (VLMs) struggle to perform this process reliably.\ Agentic systems address this challenge by querying external vision and geometry tools, but training their reasoning leaves these tools unchanged, so inaccurate observations can still lead to incorrect answers.\ We introduce \methodname{}, a framework that internalizes a plan-perceive-compose process within a single VLM. To this end, we first train specialized Planner and Perception teachers using separate adapters (Dual-LoRA), then fine-tune the corresponding experts in a mixture-of-experts (MoE) student under their supervision.\ We subsequently train the student on trajectories generated by its current policy using a segment-level extension of multi-teacher on-policy distillation (MOPD), assigning different trajectory segments to different supervisors. The Planner teacher supervises which evidence-seeking questions to ask, when to stop gathering evidence, and how to compose the final answer. The Perception teacher supervises the video-grounded answers to those questions. At inference, the student executes this adaptive workflow end-to-end, without external spatial tools. Across three spatial reasoning benchmarks, \methodname{} students built from Qwen3-VL-8B and SenseNova-SI-1.3-8B outperform their base models by an average of 9.2 and 7.5 percentage points, respectively. Across both backbones, \methodname{} reduces inference FLOPs by 65.8-66.3\% and achieves $1.29$-$1.49\times$ speedups over their shared Dual-LoRA teacher.

\end{abstract}%
  \endgroup
}

\begin{document}
\maketitle

\section{Introduction}

\begin{figure}[h]
    \centering
    \includegraphics[width=\linewidth]{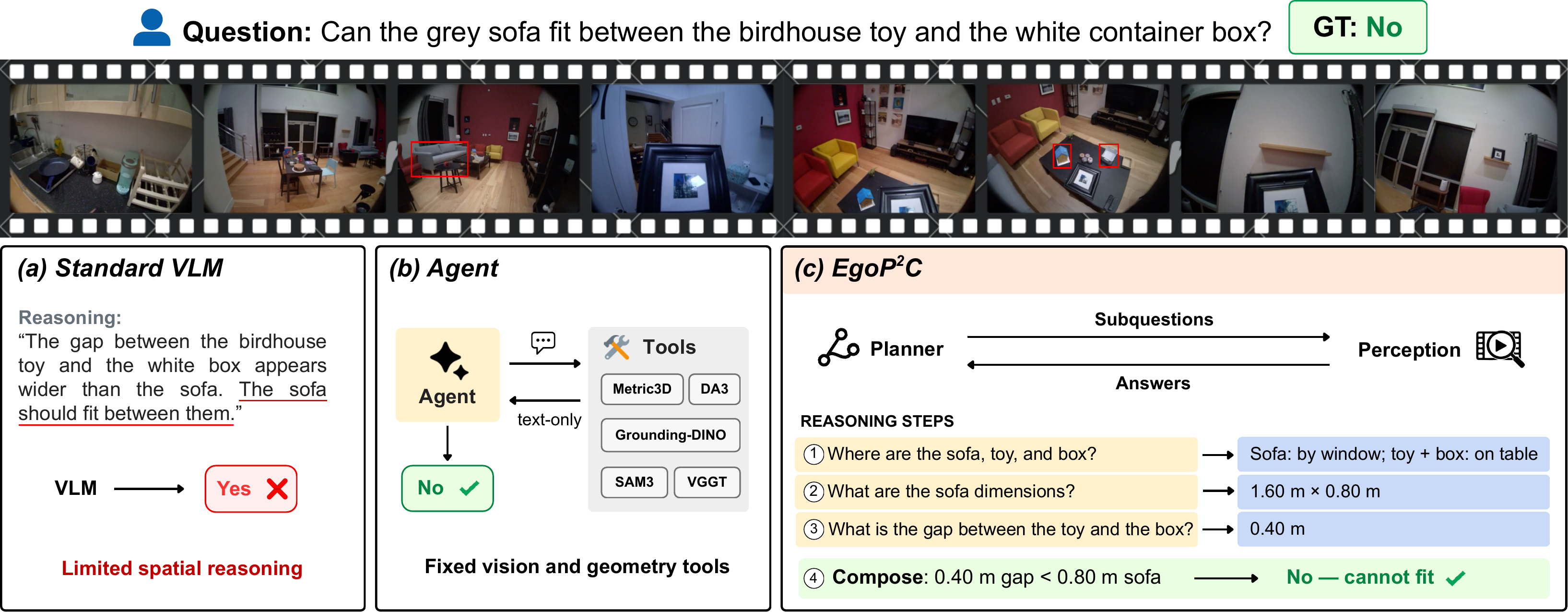}
\caption{\textbf{\methodname{} internalizes planning, perception, and composition within a single VLM for egocentric spatial reasoning.}
(a) A standard VLM reasons directly from the video and incorrectly concludes that the sofa fits between the objects.
(b) An agentic system queries external vision and geometry tools to obtain the required measurements.
(c) In \methodname{}, the Planner generates targeted questions and decides when enough evidence has been gathered, Perception answers those questions from the video, and the Planner composes the final answer.}

    \label{fig:intro_fig}
    \vspace{-1mm}
\end{figure}

Spatial reasoning in egocentric (first-person) video requires understanding object locations, distances, and spatial constraints~\citep{li2026egoprox,horton2026r3d}. This capability can support  wearable assistants~\citep{majumdar2024openeqa, huang2025vinci}, visual assistance for blind and low vision users~\citep{xiao2025egoblind, liu2026objectfinder}, robot navigation and manipulation~\citep{cheng2025navila,driess2023palme}, and augmented reality~\citep{yuan2025robopoint}. However, camera motion, occlusion, and a limited field of view cause relevant objects to appear only briefly or partially, scattering evidence across frames~\citep{yang2025thinking,horton2026r3d}. To answer complex spatial questions, a model must determine which spatial facts are needed, extract the observations from this fragmented evidence~\citep{li2026frameoracle}, and combine them through multiple reasoning steps. For example, deciding whether the sofa fits between the two objects in Figure~\ref{fig:intro_fig} requires locating the objects, estimating the sofa dimensions and available gap, and comparing these measurements.
An incorrect spatial estimate can invalidate the answer even when the subsequent reasoning is sound.

Prior work improves spatial understanding either by training vision-language models (VLMs) on spatial questions~\citep{chen2024spatialvlm} or by providing estimated 3D object coordinates as textual input~\citep{gholami2025ego3d}.
These models reason over the supplied video inputs or estimated object coordinates, but are not explicitly trained to recognize when those inputs are insufficient and seek further evidence.
Agentic systems query external vision and geometry tools for evidence at inference~\citep{chen2025spacetools,zhang2026think3d,dai2026sagent}, but the tools remain fixed: improving the reasoning component does not improve the measurements produced by those tools.


We introduce \methodname{} (Egocentric Plan-Perceive-Compose), a framework that implements these three stages through two specialist roles: \emph{Planner} and \emph{Perception}. Given a video and a spatial question, the Planner either selects direct answering or generates subquestions for Perception and uses its answers to compose the final response. Neither path uses external spatial tools. Because training both roles with shared parameters can lead to conflicting updates, we first train separate Planner and Perception teachers using low-rank adaptation (LoRA) modules on a shared frozen VLM backbone~\citep{hu2022lora}. In this Dual-LoRA configuration, each adapter is trained through supervised fine-tuning (SFT) followed by reinforcement learning using Group Relative Policy Optimization (GRPO)~\citep{shao2024deepseekmath}.

Executing the Dual-LoRA teachers requires adapter switching and repeated context processing, so we integrate their capabilities into one student while retaining separate planning and perception parameters.\
A mixture-of-experts (MoE) architecture provides role experts within a shared backbone.\
We construct this student through sparse upcycling~\citep{komatsuzaki2023sparse}, which copies selected feedforward networks of a dense VLM into experts for planning and composition, perception, and direct answering.\
We initialize the Planner and Perception experts using their corresponding teachers' supervision.\
The student executes the interaction within one checkpoint and reuses a common video prefix across roles.\
To train the initialized student on its trajectories, we extend multi-teacher on-policy distillation (MOPD)~\citep{ma2026mopd} by assigning specialist teachers to corresponding trajectory segments.\
The Planner teacher supervises planning and composition, and the Perception teacher supervises observations.
The frozen base VLM supervises direct answers.

We evaluate \methodname{} initialized from Qwen3-VL-8B and SenseNova-SI-1.3-8B on three egocentric spatial reasoning benchmarks. Both students outperform their respective base models and reduce inference FLOPs and latency relative to the shared Dual-LoRA teacher system.


In summary, our contributions are as follows:
\begingroup
\setlength{\parskip}{0pt}
\begin{itemize}[leftmargin=*,labelindent=10pt,labelsep=0.5em, itemsep=0pt,parsep=0pt,topsep=4pt,partopsep=0pt]
    \item We introduce \methodname{}, a framework that internalizes a plan-perceive-compose process within a single VLM. The resulting model answers complex egocentric spatial reasoning questions by generating and answering subquestions without external tools.
    \item We develop segment-level MOPD, assigning the Planner teacher to planning and composition segments, and the Perception teacher to observation segments within each student rollout.
    \item On three spatial reasoning benchmarks, the Qwen3-VL-8B and SenseNova-SI-1.3-8B students consistently outperform their respective base models, with average gains of 9.2 and 7.5 percentage points, respectively.\ Both students also reduce inference FLOPs by 65.8-66.3\% and achieve a $1.29$-$1.49\times$ speedup over the shared Dual-LoRA teachers.
\end{itemize}
\endgroup

\section{Related Work}

\subsection{Egocentric Spatial Reasoning}
Egocentric spatial reasoning requires inferring object relations and scene geometry from partial observations captured by a moving camera~\citep{yang2025thinking,li2026egoprox,horton2026r3d,kulkarni2026egovita}.
Existing methods train VLMs on object relations~\citep{chen2024spatialvlm}, distance estimation~\citep{cai2025sensenova}, depth understanding~\citep{cai2024spatialbot}, and camera pose estimation~\citep{cai2026vlm3}.
Others represent geometry explicitly through textual maps of estimated object coordinates~\citep{gholami2025ego3d} or descriptions of camera trajectories and scene structure~\citep{hua2026trace}.
Spatial agents obtain evidence through visual programs~\citep{marsili2025vadar,cho2026spatialclaw,luo2026pyspatial}, vision and geometry tools~\citep{chen2025spacetools,zhang2026think3d,dai2026sagent}, or world simulation~\citep{zhu2026astra}.
Recent methods also transfer spatial knowledge through synthetic viewpoint changes~\citep{zhang2026world2vlm}, depth and camera geometry supervision~\citep{tang2026spatialsv}, cognitive-map distillation~\citep{li2026omniviewspace}, and supervision from tool interactions~\citep{zhou2026spatialcli}.
\methodname{} focuses on integrating separately trained planning and perception capabilities into one VLM, with specialist teachers supervising the model's own trajectories.

\subsection{Mixture-of-Experts Models}
Sparse MoE models increase parameter capacity while limiting computation by activating a subset of experts for each token~\citep{fedus2022switch,jiang2024mixtral,dai2024deepseekmoe,muennighoff2024olmoe}.
Sparse Upcycling~\citep{komatsuzaki2023sparse} constructs an MoE from a pretrained dense checkpoint by copying its feedforward layers into experts.
Branch-Train-MiX~\citep{sukhbaatar2024btx} and FlexOlmo~\citep{shi2025flexolmo} combine experts trained on different data domains, allowing specialization before integration.
BAR~\citep{morrison2026bar} extends modular expert training to post-training, combining domain experts through parameter merging and lightweight router training.
In multimodal settings, MoSE~\citep{xu2025mose} organizes experts around skills such as perception, prediction, and planning, using skill supervision to train the routers.
To integrate separately trained capabilities, \methodname{} initializes role-specific experts through supervision from specialist teachers.
Each segment activates the expert assigned to its role.

\subsection{On-Policy Distillation}
Recent distillation methods improve language and vision-language models using teacher feedback on student-generated sequences~\citep{gu2024minillm,ko2024distillm,ko2025distillm2,yuan2026visionopd}.
In on-policy distillation (OPD), this feedback is computed on sequences sampled from the current student~\citep{agarwal2024opd}.
MOPD~\citep{ma2026mopd} extends this approach to multiple specialist teachers, selecting one teacher for each student trajectory according to the task domain.
Other methods adapt teacher supervision at the token or step level~\citep{wang2026madopd,xu2026tip}.
ViGOS~\citep{wang2026vigos} supervises visual descriptions and subsequent reasoning with the same frozen model under different contexts.
H-OPD~\citep{yin2026hopd} combines vision-language and text-only teachers through confidence-weighted token-level supervision.
\methodname{} instead assigns Planner and Perception teachers to their corresponding segments in the trajectories generated by the student, using their token-level feedback for segment-level MOPD.

\section{Method}
\label{sec:method}

\begin{figure}[t]
    \centering
    \includegraphics[width=\linewidth]{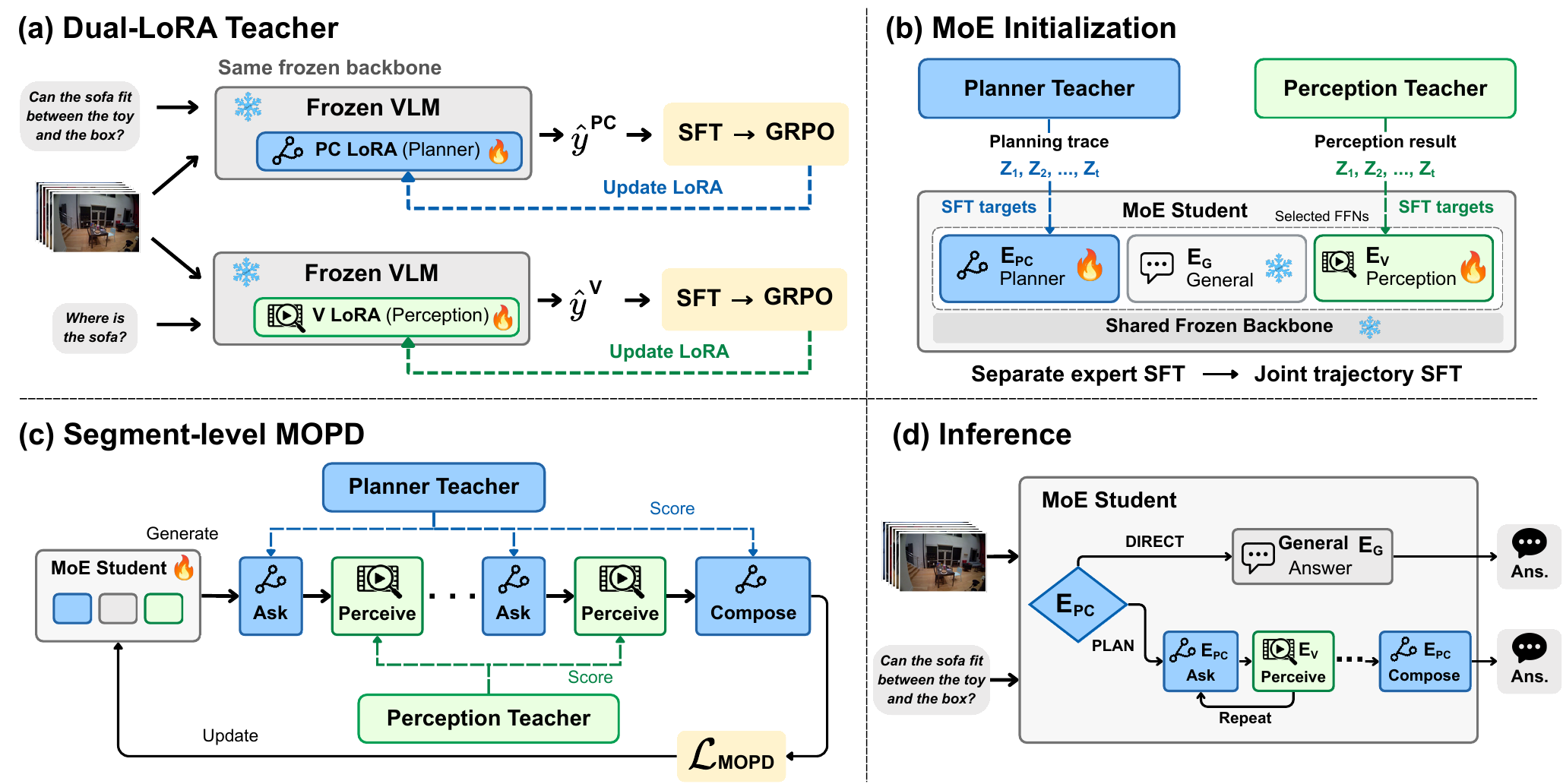}
    \caption{\textbf{Overview of \methodname{}.}
The framework integrates planning, perception, and composition into a single VLM.
(a) Planner (PC) and Perception (V) teachers share a frozen backbone and use separate LoRA adapters trained with SFT followed by GRPO.
(b) Teacher-generated sequences initialize corresponding MoE experts through separate role training followed by joint SFT.
$z_t$ denotes the $t$-th token.
(c) Segment-level MOPD trains the student on its own interactions, with PC supervising subquestions and composition and V supervising observations.
(d) At inference, the Planner expert either answers directly using the General expert (\texttt{DIRECT}) or iteratively requests subquestion answers from Perception before composing the final answer (\texttt{PLAN}).}
    \label{fig:pipeline_fig}
    \vspace{-3mm}
\end{figure}


Given an input $x{=}(v,q)$, which consists of an egocentric video $v$ and a spatial question $q$, the task is to predict the corresponding answer $a$.
Under our protocol, the question is answered either directly or by following a plan-perceive-compose process.

The Planner first selects a mode
$b\in\{\texttt{DIRECT},\texttt{PLAN}\}$.
With \texttt{DIRECT}, the model answers the original question without requesting observations.
With \texttt{PLAN}, the Planner requests spatial observations over one or more rounds before composing the answer.
At round $j$, it generates a request $u_j$ containing one or more self-contained spatial subquestions.
Perception answers each from the video, producing a list of textual observations $o_j$.
The Planner uses the accumulated requests and observations to either request further subquestions or compose the final answer. 
Perception receives only the video and its current subquestion.
Let $J$ denote the number of planning rounds and $c$ the final composition, including the answer $a$.
The output is $y{=}a$ for \texttt{DIRECT} and $y{=}(u_1,o_1,\ldots,u_J,o_J,c)$ for \texttt{PLAN}.
Appendix~\ref{app:inference_details} provides the full execution details and probabilistic formulation.

Figure~\ref{fig:pipeline_fig}(a-c) summarizes three training stages.
We train Planner and Perception teachers in a \emph{Dual-LoRA} configuration (Section~\ref{sec:role_teachers}).
These teachers provide supervision for initializing the corresponding roles in a MoE student (Section~\ref{sec:student_initialization}).
We then apply segment-level MOPD, where each teacher supervises its corresponding segments of student-generated trajectories (Section~\ref{sec:segment_distillation}).
At inference, the student's Planner expert selects the General expert for direct answering or iteratively asks the Perception expert to answer subquestions before composing the final answer (Figure~\ref{fig:pipeline_fig}(d)).

\subsection{Specialist Teachers}
\label{sec:role_teachers}

The Dual-LoRA teachers share a pretrained VLM backbone, $\pi_{\mathrm B}$ (Figure~\ref{fig:pipeline_fig}(a)).
Its vision encoder, visual-language aligner, and base language parameters remain frozen. 
Separate LoRA adapters in the language model~\citep{hu2022lora} define the Planner policy, $\pi_{\mathrm{PC}}$, for both planning and composition, and the Perception policy, $\pi_{\mathrm{V}}$.
Direct answers use $\pi_{\mathrm B}$ with both adapters disabled.

Planner SFT covers the initial \texttt{DIRECT}/\texttt{PLAN} choice.
For examples in \texttt{PLAN} mode, we also supervise subquestion generation and final answer composition.
Perception SFT covers answers to those subquestions.
For examples in \texttt{DIRECT} mode, we supervise only the Planner's mode-selection token, excluding the model's answer from the SFT loss.
During the data construction process, a language model generates subquestions and answer composition without access to the correct final answer.
Because an annotation for the original question does not necessarily provide answers to its subquestions, we obtain supervision for each subquestion from ground-truth answers when available or a strong perception model's weak labels otherwise.
The final answer is revealed only afterward to filter incorrect trajectories.
Section~\ref{sec:experimental_setup} summarizes the data sources, generation models, and sample counts.
After SFT, we apply GRPO in two stages, first to Perception and then to the Planner.
Without ground-truth subquestion answers, Perception GRPO uses the change in the frozen Planner's margin for the correct final option as a weak reward.
Both this setting and Planner GRPO use original multiple-choice questions with known correct options.

\noindent\textbf{Perception GRPO.} For each video and subquestion, the current Perception policy samples a group of answers.
A verifier scores each against the ground-truth subquestion answer when available.
Otherwise, we use \texttt{unknown} as a placeholder for a missing subquestion answer.
Utility is the change in the frozen SFT-trained Planner's margin for the correct final option when this placeholder is replaced by the sampled answer.
The video and other observations remain fixed.
This utility reward does not independently verify the subquestion answer.
The Planner evaluates composition without route selection or further questions.

\noindent\textbf{Planner GRPO.}
We freeze the trained Perception teacher and optimize the Planner starting from its initial route choice.
The frozen backbone answers under \texttt{DIRECT}, while the Planner queries Perception and composes the answer under \texttt{PLAN}.
Final correctness rewards the initial choice in both branches and the Planner's subsequent outputs in the interaction branch.
Auxiliary terms measure improvement over direct answering by $\pi_{\mathrm B}$ and penalize invalid outputs.
Base and Perception answer tokens are excluded from Planner updates.
Appendix~\ref{app:teacher_training} defines the rewards and training masks.

These teachers specialize in distinct roles, but their separate execution requires adapter switching and repeated context processing.
We next transfer their capabilities to one student.

\subsection{MoE Initialization}
\label{sec:student_initialization}

We construct the student from a pretrained dense VLM, $\pi_0$, by copying the feedforward networks in selected decoder layers into three role-specific experts (Figure~\ref{fig:pipeline_fig}(b)).
Each role $r\in\{\mathrm{PC},\mathrm V,\mathrm G\}$ has an expert $E_r$ for planning and composition, perception, or direct answering, respectively.
These roles execute the direct or plan-perceive-compose workflow in Figure~\ref{fig:pipeline_fig}(d).
$PC$ and $V$ also use role-specific attention residuals, while the shared backbone and $E_{\mathrm G}$ remain frozen.
The common video prefix uses the frozen base path and shares its KV cache across roles (Appendix~\ref{app:inference_details}).

We first train the $PC$ and $V$ experts and attention residuals separately.
Planner teacher sequences supervise $PC$ targets, while Perception teacher sequences supervise $V$ targets.
Available ground truth supervises subquestion answers and final option labels.
Let $\mathcal L_{\mathrm{seq}}^r$ and $\mathcal L_{\mathrm{gt}}^r$ denote the corresponding token cross-entropy losses. We minimize 
\begin{equation}
\label{eq:initialization_loss}
 \mathcal L_{\mathrm{init}}
 =\sum_{r\in\{\mathrm{PC},\mathrm V\}}
 w_r\left(\lambda_{\mathrm{seq}}\mathcal L_{\mathrm{seq}}^r
 +\lambda_{\mathrm{gt}}\mathcal L_{\mathrm{gt}}^r\right),
\end{equation}
where $w_r$ balances roles and the $\lambda$ coefficients weight supervision sources.
Joint SFT then applies the same objective to subquestion samples and complete trajectories from both routes.
For \texttt{DIRECT}, only $PC$'s mode choice is supervised, excluding $G$'s answer tokens.
Appendices~\ref{app:data_composition} and~\ref{app:student_training} detail sample reuse, trainable parameters, and masks.

\subsection{Segment-Level MOPD}
\label{sec:segment_distillation}

\begin{figure}[t]
    \centering
    \includegraphics[width=0.9\linewidth]{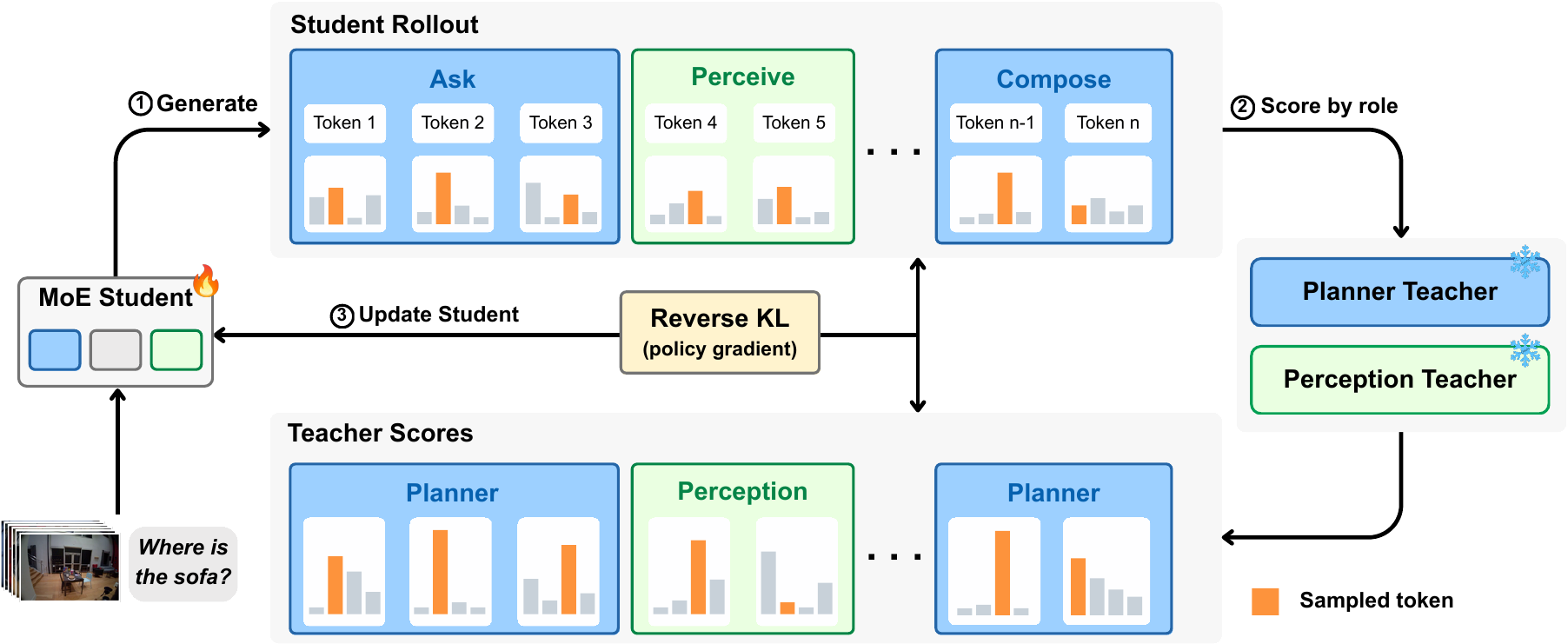}
    \caption{\textbf{Overview of the segment-level MOPD pipeline.}
Each training iteration (i) selects a training sample, (ii) generates a rollout with the MoE student model, (iii) assigns each segment to the corresponding frozen teacher, (iv) obtains teacher log probabilities for the sampled tokens, and (v) updates the student with a reverse-KL policy gradient.
The example follows the \texttt{PLAN} path.
Orange bars indicate the probabilities of the same sampled token under the student (top) and its assigned teacher (bottom).}
    \label{fig:seg_mopd_fig}
\end{figure}

We apply MOPD~\citep{ma2026mopd} to student trajectories spanning the initial mode choice through the final answer. The Planner teacher supervises the mode choice, subquestions, and composition segments.
The Perception teacher supervises observation segments, and the frozen teacher backbone $\pi_{\mathrm B}$ supervises direct answer segments.
At this stage, we introduce an attention residual for $G$ and train it with the $PC$ and $V$ parameters.
The shared backbone and $E_{\mathrm G}$ remain frozen.
Let $T(r)$ denote the teacher assigned to role $r$.
Figure~\ref{fig:seg_mopd_fig} illustrates this process for a \texttt{PLAN} trajectory.

Let $\pi_\theta^r$ denote the student policy for role $r$.
At sampling time, the student parameters $\bar\theta$ induce the trajectory distribution $p_{\bar\theta}(\cdot\mid x)$ under the interaction protocol defined above.
We sample $Y\sim p_{\bar\theta}(\cdot\mid x)$, where $Y$ contains all generated tokens.
The frozen teachers score these tokens using the same role-specific information available during trajectory generation and student updates.
For the $t$-th generated token $Y_t$, with role $r_t$ and context $C_t^{r_t}$, the clipped policy gradient advantage is
\begin{equation}
\label{eq:distillation_advantage}
 \widetilde A_t=
 \operatorname{clip}\!\left(
 \operatorname{sg}\!\left[
 \log\frac{T(r_t)(Y_t\mid C_t^{r_t})}
 {\pi_{\bar\theta}^{r_t}(Y_t\mid C_t^{r_t})}
 \right],-A_{\max},A_{\max}\right),
\end{equation}
where $\operatorname{sg}$ stops gradients and $A_{\max}$ bounds the advantage.
For segment $s\in\mathcal S(Y)$, let $r_s$ be its role and $\mathcal I_s$ its target positions.
Let $N_r(Y)$ count all supervised tokens of role $r$. We minimize
\begin{equation}
\label{eq:segment_mopd}
 \mathcal L_{\mathrm{MOPD}}(\theta)
 =-\mathbb E_{x,\,Y\sim p_{\bar\theta}}
 \left[\sum_{s\in\mathcal S(Y)}
 \frac{\omega_{r_s}}{N_{r_s}(Y)}
 \sum_{t\in\mathcal I_s}
 \widetilde A_t\log\pi_\theta^{r_s}(Y_t\mid C_t^{r_s})\right].
\end{equation}
The objective uses a clipped local reverse-KL surrogate, with $\omega_r$ balancing roles and $N_r(Y)$ normalizing their supervised token counts.
The student executes both routes within one checkpoint.
Each role selects its corresponding FFN expert and attention residual while reusing the common video KV cache.
Appendix~\ref{app:student_training} details scoring, and Appendix~\ref{app:inference_details} describes execution and cache reuse.
\section{Experiments}
\label{sec:experiments}

We evaluate \methodname{} on egocentric spatial reasoning benchmarks and analyze inference efficiency, teacher specialization, and the effects of MoE initialization and segment-level MOPD.

\subsection{Experimental setup}
\label{sec:implementation}
\label{sec:experimental_setup}

\noindent\textbf{Implementation.} The Planner and Perception teachers share the SenseNova-SI-1.3-Qwen3-VL-8B backbone~\citep{cai2025sensenova}, with separate rank-16 LoRA modules in the language model.
Students use Qwen3-VL-8B~\citep{bai2025qwen3vl} or SenseNova-SI-1.3-Qwen3-VL-8B as the backbone, abbreviated as Qwen3-VL and SenseNova-SI, respectively.
Both student models use 12 MoE layers and rank-16 attention residuals.
Student SFT runs for three epochs in each of the training stages with a learning rate of $5 \times 10^{-5}$ and a batch size of 16.
MOPD runs for a total of 600 updates with a learning rate of $3 \times 10^{-5}$ and 16 trajectories per update.
We conduct training on eight NVIDIA H100 GPUs.
Appendix~\ref{app:implementation} provides additional implementation details on model parameterization, training objectives, and configurations.

\noindent\textbf{Training Data.}
We construct training samples from Aria Digital Twin (ADT)~\citep{pan2023adt}, Ego-Exo4D~\citep{grauman2024egoexo4d}, VSI-590K~\citep{yang2025cambrians}, SPAR-7M~\citep{zhang2025spar}, and Ego4D~\citep{grauman2022ego4d}.
The videos used for model training are disjoint from those included in the evaluation benchmarks.
For teacher SFT, the DeepSeek-V4-Pro API~\citep{xu2026deepseek} generates subquestions and composes the final answer.
Subquestion answers come from ground-truth annotations when available; otherwise, they come from the Qwen3.5-397B API~\citep{qwen2026qwen35}.
The Perception and Planner teachers use $35,811$ and $17,817$ samples for SFT, and $3,160$ and $3,872$ samples for GRPO, respectively.
Student initialization draws $15,000$ Perception and $4,100$ Planner samples from these pools, followed by joint SFT on 8,000 samples.
MOPD uses a total of 8,000 prompts.
Appendices~\ref{app:data_construction} and~\ref{app:prompts} provide construction details, sample distributions, and prompt templates.


\noindent\textbf{Benchmarks.}\ We evaluate \methodname{} on three egocentric spatial reasoning benchmarks.\ 
VSI-Bench~\citep{yang2025thinking} covers spatial relations, distance estimation, and route planning in indoor scenes.
EgoProx~\citep{li2026egoprox} tests egocentric navigation and object interactions using spatial relations and action context. We use its single-answer multiple-choice tasks and exclude Chain of Actions, which requires structured action-sequence prediction.
R3D-Bench~\citep{horton2026r3d} tests spatial reasoning from wearable video, including comparing and combining object distance and dimension measurements.
We do not provide models with R3D-Bench’s depth maps or camera poses. Agentic baselines may estimate them from RGB inputs using external tools.


\noindent\textbf{Evaluation Settings.}
We sample videos at 1 fps with a maximum of 32 frames per input, and cap each model generation at 2,048 tokens.
Dual-LoRA and \methodname{} may ask at most six Perception subquestions per question.
The same sampled frames are used within each controlled comparison.
For the S-Agent variants, we keep the external tools, prompts, and execution settings fixed and vary only the agent.

\begin{table*}[t]
\centering
\caption{\textbf{Performance on egocentric spatial reasoning benchmarks.}
Subscripts indicate the backbone or agent model.
\textcolor{darkgreen}{Green parentheses} report gains in percentage points over the baseline row.
Student gains on all three benchmarks have 95\% bootstrap confidence intervals excluding zero (Appendix~\ref{app:statistical_analysis}).
}
\label{tab:main_results}
\small
\setlength{\tabcolsep}{6pt}
\resizebox{0.86\textwidth}{!}{%
\renewcommand{\arraystretch}{1.1}
\begin{tabular}{c l | c c c | c}
\toprule
\textbf{\#} &
\textbf{Method} &
\textbf{VSI-Bench} &
\textbf{EgoProx} &
\textbf{R3D-Bench} &
\textbf{Avg.} \\
\midrule

\rowcolor{gray!15}
\multicolumn{6}{c}{\textit{Agentic Systems (w/ tools)}} \\
\midrule
1 & SpatialClaw\textsubscript{\textbf{ Gemma4-31B}}~\citep{cho2026spatialclaw}
& 48.5 & 47.9 & 40.7 & 45.7 \\
2 & Think3D\textsubscript{\textbf{ Gemini 2.5 Pro}}~\citep{zhang2026think3d}
& 62.0 & 42.1 & 43.4 & 49.2 \\
3 & S-Agent\textsubscript{\textbf{ Qwen3-VL-8B}}~\citep{dai2026sagent}
& 56.3 & 30.2 & 33.9 & 40.1 \\
4 & S-Agent\textsubscript{\textbf{ Qwen3-VL-30B-A3B}}~\citep{dai2026sagent}
& 64.8 & 42.7 & 41.4 & 49.6 \\
5 & S-Agent\textsubscript{\textbf{ GPT-5.4}}~\citep{dai2026sagent}
& 66.7 & 49.8 & 45.9 & 54.1 \\

\midrule
\rowcolor{gray!15}
\multicolumn{6}{c}{\textit{Open-source General Models (w/o tools)}} \\
\midrule
6 & Qwen3-VL-8B~\citep{bai2025qwen3vl}
& 57.9 & 31.0 & 35.2 & 41.4 \\
7 & Qwen3.5-9B~\citep{qwen2026qwen35}
& 54.3 & 32.8 & 37.3 & 41.5 \\
8 & InternVL3.5-8B~\citep{wang2025internvl35}
& 56.1 & 29.3 & 33.6 & 39.7 \\
9 & Qwen3-VL-30B-A3B~\citep{bai2025qwen3vl}
& 63.2 & 38.5 & 39.0 & 46.9 \\

\midrule
\rowcolor{gray!15}
\multicolumn{6}{c}{\textit{Open-source Spatial Models (w/o tools)}} \\
\midrule
10 & VST-7B-SFT~\citep{yang2026vst}
& 60.6 & 29.2 & 31.4 & 40.4 \\
11 & Cambrian-S-7B~\citep{yang2025cambrians}
& 62.9 & 27.1 & 36.3 & 42.1 \\
12 & SenseNova-SI-1.3-8B~\citep{cai2025sensenova}
& 64.8 & 34.2 & 37.1 & 45.4 \\
13 & RynnBrain-8B~\citep{dang2026rynnbrain}
& 68.6 & 39.6 & 35.9 & 48.0 \\

\midrule
\rowcolor{gray!15}
\multicolumn{6}{c}{\textit{Proprietary Models (w/o tools)}} \\
\midrule
14 & Gemini 2.5 Pro~\citep{comanici2025gemini}
& 53.5 & 40.3 & 39.5 & 44.4 \\
15 & Gemini 3 Pro~\citep{pichai2025gemini3}
& 53.7 & 42.9 & 44.3 & 47.0 \\
16 & GPT-5.5~\citep{openai2026gpt55}
& 60.4 & 44.2 & 44.7 & 49.8 \\

\midrule
\rowcolor{gray!15}
\multicolumn{6}{c}{\textit{Ours (w/o tools)}} \\
\midrule
\rowcolor{yellow}
17 & Dual-LoRA\textsubscript{\textbf{ SenseNova-SI}}
& 67.6 & 49.0 & 44.8 & 53.8 \\

\rowcolor{blue!8}
18 & \textbf{\methodname{}\textsubscript{\textbf{ Qwen3-VL} {\scriptsize\textcolor{darkgreen}{($\Delta$ vs.\ \#6)}}}}
& \textbf{62.1 {\scriptsize\textcolor{darkgreen}{(+4.2)}}}
& \textbf{45.9 {\scriptsize\textcolor{darkgreen}{(+14.9)}}}
& \textbf{43.7 {\scriptsize\textcolor{darkgreen}{(+8.5)}}}
& \textbf{50.6 {\scriptsize\textcolor{darkgreen}{(+9.2)}}} \\

\rowcolor{blue!8}
19 & \textbf{\methodname{}\textsubscript{\textbf{ SenseNova-SI} {\scriptsize\textcolor{darkgreen}{($\Delta$ vs.\ \#12)}}}}
& \textbf{66.9 {\scriptsize\textcolor{darkgreen}{(+2.1)}}}
& \textbf{47.2 {\scriptsize\textcolor{darkgreen}{(+13.0)}}}
& \textbf{44.5 {\scriptsize\textcolor{darkgreen}{(+7.4)}}}
& \textbf{52.9 {\scriptsize\textcolor{darkgreen}{(+7.5)}}} \\
\bottomrule
\end{tabular}%
}
\vspace{-3mm}
\end{table*}

\begin{table}[t]
\centering
\caption{\textbf{Inference cost of Dual-LoRA and \methodname{}.}
Subscripts identify the starting checkpoints.
\textcolor{darkgreen}{Green parentheses} report relative reductions from Dual-LoRA.
FLOPs, latency, and generated tokens are averaged over all questions across the three benchmarks.
Generated tokens include outputs from all roles.
}
\label{tab:inference_efficiency}
\small
\setlength{\tabcolsep}{5pt}
\renewcommand{\arraystretch}{1.1}
\resizebox{0.81\linewidth}{!}{%
\begin{tabular}{lcccc}
\toprule
\textbf{Method} &
\textbf{TFLOPs} &
\textbf{Latency (s)} &
\textbf{Generated tokens} &
\textbf{Latency/token (ms)} \\
\midrule
Dual-LoRA\textsubscript{\textbf{ SenseNova-SI}}
& 165.7
& 9.4
& 353.1
& 26.6 \\
\rowcolor{blue!8}
\textbf{\methodname{}}\textsubscript{\textbf{ Qwen3-VL}}
& \textbf{55.9 $\scriptstyle\textcolor{darkgreen}{\textbf{(-66.3\%)}}$}
& \textbf{7.3 $\scriptstyle\textcolor{darkgreen}{\textbf{(-22.3\%)}}$}
& \textbf{315.4 $\scriptstyle\textcolor{darkgreen}{\textbf{(-10.7\%)}}$}
& \textbf{23.1 $\scriptstyle\textcolor{darkgreen}{\textbf{(-13.2\%)}}$} \\
\rowcolor{blue!8}
\textbf{\methodname{}}\textsubscript{\textbf{ SenseNova-SI}}
& \textbf{56.6 $\scriptstyle\textcolor{darkgreen}{\textbf{(-65.8\%)}}$}
& \textbf{6.3 $\scriptstyle\textcolor{darkgreen}{\textbf{(-33.0\%)}}$}
& \textbf{276.8 $\scriptstyle\textcolor{darkgreen}{\textbf{(-21.6\%)}}$}
& \textbf{22.8 $\scriptstyle\textcolor{darkgreen}{\textbf{(-14.3\%)}}$} \\
\bottomrule
\end{tabular}%
}
\end{table}

\subsection{Results}
\label{sec:results}

Table~\ref{tab:main_results} shows that both students significantly improve over their base models on all three benchmarks.\
\methodname{}\textsubscript{\textbf{ Qwen3-VL}} reaches an average score of 50.6, exceeding the larger Qwen3-VL-30B-A3B model at 46.9.
When used as the planner in S-Agent, Qwen3-VL-8B's average score decreases from 41.4 to 40.1, indicating that it does not benefit from the external tools under this setting.
Qwen3-VL-30B-A3B improves in average score from 46.9 to 49.6 with S-Agent, but remains below both 8B \methodname{} students.
Among the evaluated S-Agent configurations, only GPT-5.4 exceeds both \methodname{} students on average, reaching 54.1.
These comparisons show that S-Agent's effectiveness depends strongly on the capability of its planner model, and \methodname{} enables compact open-source VLMs to learn the complete plan-perceive-compose interaction.
The SenseNova-SI results further show that \methodname{} improves a backbone already trained for spatial understanding.
The resulting student preserves much of the Dual-LoRA teachers' performance while executing all roles within a single checkpoint and reusing a shared video prefix.

The subtask results reported in Appendix~\ref{app:category_results} also show that both student models outperform their respective base models on each of the evaluated subtasks.
On VSI-Bench, \methodname{}\textsubscript{\textbf{ Qwen3-VL}} and \methodname{}\textsubscript{\textbf{ SenseNova-SI}} gain 7.2 and 4.4 points on \emph{Object Absolute Distance}, which requires estimating the distance between two objects.
On R3D-Bench, they gain 12.7 and 11.8 points on \emph{Total Walk Distance}, which requires summing the horizontal distances from the camera wearer to one object and from that object to another.
On EgoProx, they gain 29.3 and 24.1 points on \emph{Exploration Relative}, which asks for the direction of the next navigation step toward a goal.
These results show that \methodname{} improves basic spatial perception and achieves larger gains on tasks that combine multiple spatial observations.

To test whether additional QA training alone accounts for the reported performance gains, we train a SenseNova-SI baseline model with SFT + GRPO on the same original questions and subquestion QA data as the Dual-LoRA teacher.
During the SFT stage, the baseline is trained only on answers and receives no supervision on intermediate planning and interaction sequences.
GRPO uses both the Perception training samples and the original multiple-choice questions used for Planner GRPO.
At inference time, the baseline answers directly, without generating subquestions.
Appendix~\ref{app:direct_answer_baseline} shows statistically significant performance gains for both Dual-LoRA and \methodname{}\textsubscript{\textbf{ SenseNova-SI}} over this baseline.
\methodname{}\textsubscript{\textbf{ SenseNova-SI}} improves by $1.4$, $6.5$, and $4.4$ points on VSI-Bench, EgoProx, and R3D-Bench, respectively.
These results support that the value of plan-perceive-compose goes beyond directly learning answers from the same QA data.


\noindent\textbf{Inference Efficiency.} We compare both \methodname{} students with their shared SenseNova-SI Dual-LoRA teachers across all three benchmarks.
All models use vLLM~\citep{kwon2023vllm} with the same hardware, video sampling, resolution, batch size, and generation limits.
Each system follows its own routing and stopping decisions.
Measurements cover the complete interaction, including visual encoding and all generation steps. 
Table~\ref{tab:inference_efficiency} shows that \methodname{}\textsubscript{\textbf{ Qwen3-VL}} and \methodname{}\textsubscript{\textbf{ SenseNova-SI}} reduce FLOPs by 66.3\% and 65.8\% and achieve $1.29\times$ and $1.49\times$ speedups over Dual-LoRA, respectively.
Both students generate fewer tokens and have lower end-to-end latency per generated token, despite making slightly more Perception calls (2.6 and 2.7 vs. 2.5 per question).
Peak GPU memory increases from 19.4 GB for Dual-LoRA to 25.2 and 25.3 GB for the two students. 
The efficiency gains also persist when comparing subsets in which each student and Dual-LoRA generate the same number of subquestions for each original question (Appendix~\ref{app:efficiency}).

\subsection{Ablation Studies}

\begin{table}[t]
\centering
\caption{\textbf{Further supervised training of the student.}
All systems use the SenseNova-SI-1.3-8B backbone.
Average is the unweighted mean across the three benchmarks.}
\label{tab:continued_sft}
\small
\setlength{\tabcolsep}{5pt}
\resizebox{0.67\linewidth}{!}{%
\renewcommand{\arraystretch}{1.1}
\begin{tabular}{lcccc}
\toprule
\textbf{Method} &
\textbf{VSI-Bench} &
\textbf{EgoProx} &
\textbf{R3D-Bench} &
\textbf{Avg.} \\
\midrule
Dual-LoRA teachers
& \textbf{67.6} & 49.0 & 44.8 & 53.8 \\
\methodname{}
& 66.9 & 47.2 & 44.5 & 52.9 \\
\rowcolor{blue!8}
\methodname{} $+$ additional SFT
& 67.5 & \textbf{49.8} & \textbf{46.4} & \textbf{54.6} \\
\bottomrule
\end{tabular}
}
\end{table}

\begin{table}[t]
\centering
\caption{\textbf{Effect of joint and separate role training.}
All variants use SenseNova-SI-1.3-8B.}

\label{tab:role_training}
\small
\setlength{\tabcolsep}{5pt}
\resizebox{0.68\textwidth}{!}{%
\renewcommand{\arraystretch}{1.1}
\begin{tabular}{llccc}
\toprule
\textbf{Method} &
\textbf{Training} &
\textbf{VSI-Bench} &
\textbf{EgoProx} &
\textbf{R3D-Bench} \\
\midrule
Base
& --
& 64.8 & 34.2 & 37.1 \\
\midrule
Joint LoRA
& SFT
& 62.1 & 37.9 & 37.7 \\
Perception-only LoRA
& SFT
& 65.4
& 39.0
& 39.2 \\
Planner-only LoRA
& SFT
& 61.8
& 43.9
& 41.0 \\
Dual-LoRA
& SFT
& 65.8
& 46.7
& 43.2 \\
\midrule
Joint LoRA
& SFT + GRPO
& 62.6 & 36.3 & 37.3 \\
\rowcolor{blue!8}
\textbf{Dual-LoRA}
& SFT + GRPO
& \textbf{67.6} & \textbf{49.0} & \textbf{44.8} \\
\bottomrule
\end{tabular}
}
\end{table}

\noindent\textbf{Further Supervised Training of the Student.}
To assess improvement after MOPD, we apply additional SFT to \methodname{}\textsubscript{\textbf{ SenseNova-SI}} using 3,000 samples in total, drawn from the Planner and Perception SFT pools.
Although these samples were used to train the teachers, they were not used for training the student in its preceding training stages.
We update the Planner and Perception experts without retraining the teachers or repeating MOPD.
Table~\ref{tab:continued_sft} shows that additional SFT improves the student's scores on all three benchmarks, raising its average from 52.9 to 54.6, above the Dual-LoRA teachers' 53.8.
It also surpasses the teachers on EgoProx and R3D-Bench.
These results show that the student benefits from further direct supervision using samples from the teachers’ SFT pools.

\noindent\textbf{Joint versus Separate Training of Planning and Perception.} We compare joint and separate role training using SenseNova-SI.
Joint LoRA uses a single rank-32 adapter for both roles, matching Dual-LoRA's total number of trainable adapter parameters.
We also evaluate single-role adapters trained with SFT on the corresponding role's data.
Perception-only uses its trained adapter to answer the original question without planning, while Planner-only uses the frozen base model for perception. 
Table~\ref{tab:role_training} shows that Joint SFT improves over the base model on EgoProx and R3D-Bench but reduces VSI-Bench performance.
Perception-only SFT improves over the base model on all three benchmarks, while Planner-only SFT outperforms Joint SFT on EgoProx and R3D-Bench.
After SFT, Dual-LoRA outperforms Joint LoRA and both single-role variants on all three benchmarks.
These results show that each type of supervision is useful on its own, but Joint LoRA does not consistently outperform the single-role variants.
GRPO further improves Dual-LoRA on all three benchmarks, and Joint LoRA improves slightly on VSI-Bench but declines on EgoProx and R3D-Bench.
After SFT and GRPO, Dual-LoRA outperforms Joint LoRA by 5.0, 12.7, and 7.5 points on VSI-Bench, EgoProx, and R3D-Bench, respectively. These results support using separate parameters for planning and perception while combining their capabilities during interaction.

\begin{table}[t]
\centering
\caption{\textbf{Effects of student architecture, role initialization, and distillation.}
Role initialization includes separate role SFT followed by joint SFT.
Dense $+$ SFT $+$ MOPD matches the SFT data used for role initialization.}
\label{tab:kd_initialization}
\small
\setlength{\tabcolsep}{6pt}
\resizebox{0.72\linewidth}{!}{%
\renewcommand{\arraystretch}{1.1}
\begin{tabular}{lccc}
\toprule
\textbf{Configuration} &
\textbf{VSI-Bench} &
\textbf{EgoProx} &
\textbf{R3D-Bench} \\
\midrule
\textbf{Qwen3-VL-8B}
& 57.9 & 31.0 & 35.2 \\
\quad Dense $+$ MOPD
& 58.4 & 36.3 & 32.8 \\
\quad Dense $+$ SFT $+$ MOPD
& {59.8}
& {38.6}
& {35.1} \\
\quad MoE $+$ MOPD
& 60.4 & 40.4 & 39.6 \\
\quad MoE $+$ role initialization
& 60.8 & 42.7 & 41.0 \\
\quad MoE $+$ role initialization $+$ offline distillation
& 61.3 & 43.3 & 41.2 \\
\rowcolor{blue!8}
\quad MoE $+$ role initialization $+$ MOPD
& \textbf{62.1} & \textbf{45.9} & \textbf{43.7} \\
\midrule
\textbf{SenseNova-SI-1.3-8B}
& 64.8 & 34.2 & 37.1 \\
\quad Dense $+$ MOPD
& 64.7 & 39.5 & 39.7 \\
\quad Dense $+$ SFT $+$ MOPD
& 65.2
& 43.0
& 41.9 \\
\quad MoE $+$ MOPD
& 65.3 & 44.1 & 42.4 \\
\quad MoE $+$ role initialization
& 66.0 & 45.2 & 43.3 \\
\quad MoE $+$ role initialization $+$ offline distillation
& 65.9 & 45.4 & 43.0 \\
\rowcolor{blue!8}
\quad MoE $+$ role initialization $+$ MOPD
& \textbf{66.9} & \textbf{47.2} & \textbf{44.5} \\
\bottomrule
\end{tabular}
}
\end{table}

\begin{figure*}[t]
\centering
\includegraphics[width=0.90\textwidth]{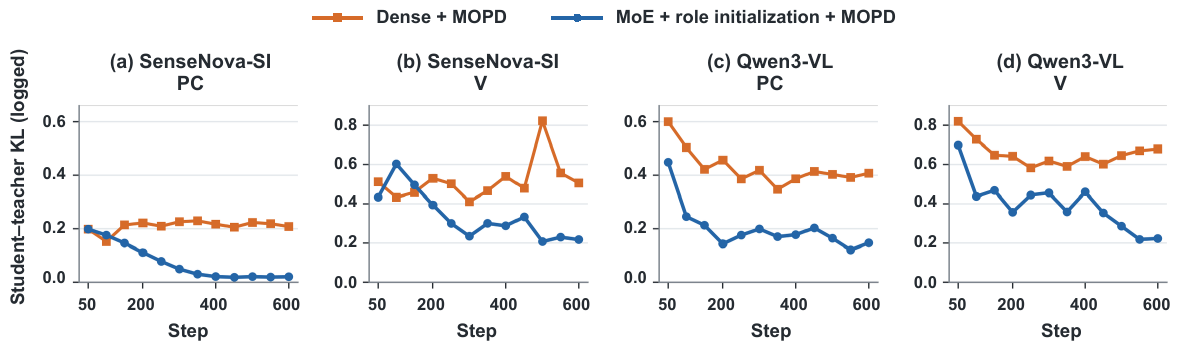}
\caption{\textbf{Student-teacher KL during MOPD.}
PC denotes planning and composition, and V denotes perception.
The initialized MoE finishes with lower KL for both roles on both backbones.}
\label{fig:role_kl}
\end{figure*}


\noindent\textbf{Effects of MoE Architecture, Role Initialization, and Distillation.} We evaluate the student design by comparing segment-level MOPD applied directly to the dense checkpoints with MOPD after MoE construction and role initialization.
We also evaluate Dense $+$ SFT $+$ MOPD, which learns Planner and Perception targets through shared dense parameters before distillation, using the same SFT data and sample exposure as MoE initialization.
Table~\ref{tab:kd_initialization} shows that adding SFT improves dense students from both backbones over Dense $+$ MOPD on all three benchmarks, but the full MoE student consistently outperforms both dense baselines.
MoE $+$ MOPD also outperforms Dense $+$ MOPD on all three benchmarks for both backbones.
Adding role initialization before MOPD further improves all three scores.
Starting from the role-initialized MoE, subsequent MOPD also provides gains on all three benchmarks.
On R3D-Bench, the full model exceeds Dense $+$ SFT $+$ MOPD by 8.6 points for Qwen3-VL and 2.6 points for SenseNova-SI.
These results show that the MoE advantage cannot be explained solely by additional SFT data exposure.

We further compare MOPD with offline distillation starting from the same role-initialized MoE students and using matched prompts and update budgets.
Offline distillation trains on fixed trajectories generated by the Dual-LoRA teachers, whereas MOPD trains on trajectories generated by the current student.
MOPD outperforms offline distillation by 0.8, 2.6, and 2.5 points for Qwen3-VL and by 1.0, 1.8, and 1.5 points for SenseNova-SI on VSI-Bench, EgoProx, and R3D-Bench, respectively.
Under this matched setting, training on the student's own trajectories is more effective than distilling fixed teacher-generated trajectories.

Beyond these benchmark results, the initialized MoE shows closer agreement with both teachers, reaching lower final KL than Dense $+$ MOPD for PC and V on both backbones (Figure~\ref{fig:role_kl}).
For SenseNova-SI, PC KL finishes at 0.021, compared with 0.208 for Dense $+$ MOPD.
Supplementary gradient analysis in Appendix~\ref{app:student_diagnostics} also finds opposing gradient directions between the two role objectives in the dense Qwen3-VL student.
Together, these diagnostics support the complete MoE initialization stage as an effective way to integrate specialist capabilities before MOPD.
\section{Conclusion}

We presented \methodname{}, a framework that learns planning and composition separately from perception and integrates these capabilities into one VLM for egocentric spatial reasoning.
Dual-LoRA teachers guide the initialization of dedicated MoE experts.
Segment-level MOPD then assigns each specialist to the corresponding segments of the student's own interactions, supervising both the observations it produces and the reasoning that uses them.
Experiments show consistent improvements over both general and spatially trained backbones. Future work will explore adaptive frame selection for longer videos and training objectives that balance answer accuracy with inference cost.

\bibliographystyle{assets/plainnat}
\bibliography{paper}

@inproceedings{hu2022lora,
  title     = {{LoRA}: Low-Rank Adaptation of Large Language Models},
  author    = {Hu, Edward J. and Shen, Yelong and Wallis, Phillip and Allen-Zhu, Zeyuan and Li, Yuanzhi and Wang, Shean and Wang, Lu and Chen, Weizhu},
  booktitle = {Proceedings of the International Conference on Learning Representations (ICLR)},
  year      = {2022}
}

@inproceedings{komatsuzaki2023sparse,
  title     = {Sparse Upcycling: Training Mixture-of-Experts from Dense Checkpoints},
  author    = {Komatsuzaki, Aran and Puigcerver, Joan and Lee-Thorp, James and Riquelme Ruiz, Carlos and Mustafa, Basil and Ainslie, Joshua and Tay, Yi and Dehghani, Mostafa and Houlsby, Neil},
  booktitle = {Proceedings of the International Conference on Learning Representations (ICLR)},
  year      = {2023}
}

@inproceedings{agarwal2024opd,
  title     = {On-Policy Distillation of Language Models: Learning from Self-Generated Mistakes},
  author    = {Agarwal, Rishabh and Vieillard, Nino and Zhou, Yongchao and Stanczyk, Piotr and Ramos, Sabela and Geist, Matthieu and Bachem, Olivier},
  booktitle = {Proceedings of the International Conference on Learning Representations (ICLR)},
  year      = {2024}
}

@article{shao2024deepseekmath,
  title   = {{DeepSeekMath}: Pushing the Limits of Mathematical Reasoning in Open Language Models},
  author  = {Shao, Zhihong and Wang, Peiyi and Zhu, Qihao and Xu, Runxin and Song, Junxiao and Bi, Xiao and Zhang, Haowei and Zhang, Mingchuan and Li, Y. K. and Wu, Y. and Guo, Daya},
  journal = {arXiv preprint arXiv:2402.03300},
  year    = {2024}
}

@article{bai2025qwen3vl,
  title   = {{Qwen3-VL} Technical Report},
  author={Bai, Shuai and Cai, Yuxuan and Chen, Ruizhe and Chen, Keqin and Chen, Xionghui and Cheng, Zesen and Deng, Lianghao and Ding, Wei and Gao, Chang and Ge, Chunjiang and others},
  journal = {arXiv preprint arXiv:2511.21631},
  year    = {2025}
}

@inproceedings{yang2025thinking,
  title     = {Thinking in Space: How Multimodal Large Language Models See, Remember, and Recall Spaces},
  author    = {Yang, Jihan and Yang, Shusheng and Gupta, Anjali W. and Han, Rilyn and Fei-Fei, Li and Xie, Saining},
  booktitle = {Proceedings of the IEEE/CVF Conference on Computer Vision and Pattern Recognition (CVPR)},
  year      = {2025}
}

@inproceedings{gholami2025ego3d,
  title     = {Spatial Reasoning with Vision-Language Models in Ego-Centric Multi-View Scenes},
  author    = {Gholami, Mohsen and Rezaei, Ahmad and Zhou, Weimin and Zhang, Yong and Akbari, Mohammad},
  booktitle = {Proceedings of the International Conference on Learning Representations (ICLR)},
  year      = {2026}
}

@inproceedings{hua2026trace,
  title     = {Unleashing Spatial Reasoning in Multimodal Large Language Models via Textual Representation Guided Reasoning},
  author    = {Hua, Jiacheng and Yin, Yishu and Wu, Yuhang and Wang, Tai and Huang, Yifei and Liu, Miao},
  booktitle = {Proceedings of the Annual Meeting of the Association for Computational Linguistics (ACL)},
  year      = {2026}
}

@inproceedings{cai2025sensenova,
  title     = {Scaling Spatial Intelligence with Multimodal Foundation Models},
  author    = {Cai, Zhongang and Wang, Ruisi and Gu, Chenyang and Pu, Fanyi and Xu, Junxiang and Wang, Yubo and Yin, Wanqi and Yang, Zhitao and Wei, Chen and Sun, Qingping and Zhou, Tongxi and Li, Jiaqi and Pang, Hui En and Qian, Oscar and Wei, Yukun and Lin, Zhiqian and Shi, Xuanke and Deng, Kewang and Han, Xiaoyang and Chen, Zukai and Fan, Xiangyu and Deng, Hanming and Lu, Lewei and Pan, Liang and Li, Bo and Liu, Ziwei and Wang, Quan and Lin, Dahua and Yang, Lei},
  booktitle = {Proceedings of the IEEE/CVF Conference on Computer Vision and Pattern Recognition (CVPR)},
  year      = {2026}
}

@article{cai2026vlm3,
  title   = {{VLM3}: Vision Language Models Are Native {3D} Learners},
  author  = {Cai, Zhipeng and Liu, Zhuang and Xiong, Yunyang and Liu, Zechun and Chandra, Vikas and Shi, Yangyang},
  journal = {arXiv preprint arXiv:2605.30561},
  year    = {2026}
}

@inproceedings{yang2026vst,
  title     = {Visual Spatial Tuning},
  author    = {Yang, Rui and Zhu, Ziyu and Li, Yanwei and Huang, Jingjia and Yan, Shen and Zhou, Siyuan and Liu, Zhe and Li, Xiangtai and Li, Shuangye and Wang, Wenqian and Lin, Yi and Zhao, Hengshuang},
  booktitle = {Proceedings of the European Conference on Computer Vision (ECCV)},
  year      = {2026}
}

@article{yang2025cambrians,
  title   = {{Cambrian-S}: Towards Spatial Supersensing in Video},
  author  = {Yang, Shusheng and Yang, Jihan and Huang, Pinzhi and Brown, Ellis and Yang, Zihao and Yu, Yue and Tong, Shengbang and Zheng, Zihan and Xu, Yifan and Wang, Muhan and Lu, Daohan and Fergus, Rob and LeCun, Yann and Fei-Fei, Li and Xie, Saining},
  journal = {arXiv preprint arXiv:2511.04670},
  year    = {2025}
}

@inproceedings{li2026egoprox,
  title     = {{EgoProx}: Evaluating {MLLMs} on Egocentric {3D} Proximity Reasoning Across a Cognitive Hierarchy},
  author    = {Li, Jinzhao and Chen, Yinuo and Piao, Dongxu and Pan, Panwang and Yu, Yifan and Wang, Dong and Yan, Honglei and Yue, Liang and Wang, Shaofei and Chen, Yixin and Huang, Siyuan and Liu, Miao},
  booktitle = {Proceedings of the IEEE/CVF Conference on Computer Vision and Pattern Recognition (CVPR)},
  year      = {2026}
}

@inproceedings{marsili2025vadar,
  title     = {Visual Agentic {AI} for Spatial Reasoning with a Dynamic {API}},
  author    = {Marsili, Damiano and Agrawal, Rohun and Yue, Yisong and Gkioxari, Georgia},
  booktitle = {Proceedings of the IEEE/CVF Conference on Computer Vision and Pattern Recognition (CVPR)},
  year      = {2025}
}

@article{cho2026spatialclaw,
  title   = {{SpatialClaw}: Rethinking Action Interface for Agentic Spatial Reasoning},
  author  = {Cho, Seokju and Hachiuma, Ryo and Badki, Abhishek and Su, Hang and Lee, Byung-Kwan and Song, Chan Hee and Liu, Sifei and Radhakrishnan, Subhashree and Kim, Seungryong and Wang, Yu-Chiang Frank and Chen, Min-Hung},
  journal = {arXiv preprint arXiv:2606.13673},
  year    = {2026}
}

@inproceedings{chen2025spacetools,
  title     = {{SpaceTools}: Tool-Augmented Spatial Reasoning via Double Interactive {RL}},
  author    = {Chen, Siyi and Uy, Mikaela Angelina and Song, Chan Hee and Ladhak, Faisal and Murali, Adithyavairavan and Qu, Qing and Birchfield, Stan and Blukis, Valts and Tremblay, Jonathan},
  booktitle = {Proceedings of the IEEE/CVF Conference on Computer Vision and Pattern Recognition (CVPR)},
  year      = {2026}
}

@article{zhang2026think3d,
  title   = {{Think3D}: Thinking with Space for Spatial Reasoning},
  author  = {Zhang, Zaibin and Wu, Yuhan and Jia, Lianjie and Wang, Yifan and Zhang, Zhongbo and Li, Yijiang and Ran, Binghao and Zhang, Fuxi and Sun, Zhuohan and Peng, Yizhuang and Yin, Zhenfei and Wang, Lijun and Lu, Huchuan},
  journal = {arXiv preprint arXiv:2601.13029},
  year    = {2026}
}

@article{zhu2026astra,
  title   = {Thinking with Imagination: Agentic Visual Spatial Reasoning with World Simulators},
  author  = {Zhu, Chenming and Lin, Jingli and Long, Yilin and Cao, Peizhou and Wang, Tai and Pang, Jiangmiao and Liu, Xihui},
  journal = {arXiv preprint arXiv:2606.06476},
  year    = {2026}
}

@article{dai2026sagent,
  title   = {{S-Agent}: Spatial Tool-Use Elicits Reasoning for Spatial Intelligence},
  author  = {Dai, Yalun and Li, Hao and Tian, Shulin and Yao, Runmao and Dong, Yuhao and Hong, Fangzhou and Chen, Zhaoxi and Liu, Fangfu and Wang, Tao and Yap, Kim-Hui and Liu, Ziwei},
  journal = {arXiv preprint arXiv:2606.20515},
  year    = {2026}
}

@article{horton2026r3d,
  title   = {{R3D}: Quantitative {3D} Spatial Reasoning for Egocentric Wearables},
  author  = {Horton, Maxwell and Lu, Wei and Tran, Quan and Astashonok, Yury and Ahmed, Kirmani and Damavandi, Babak and Kumar, Anuj and Zhang, Xiao and Moon, Seungwhan},
  journal = {arXiv preprint arXiv:2607.02921},
  year    = {2026}
}

@article{zhang2026world2vlm,
  title   = {{World2VLM}: Distilling World Model Imagination into {VLMs} for Dynamic Spatial Reasoning},
  author  = {Zhang, Wanyue and Wu, Wenxiang and Xu, Wang and Luo, Jiaxin and Zhi, Helu and Huang, Yibin and Ren, Shuo and Liu, Zitao and Zhang, Jiajun},
  journal = {arXiv preprint arXiv:2604.26934},
  year    = {2026}
}

@inproceedings{tang2026spatialsv,
  title     = {{SpatialSV}: Internalizing Interpretable {3D} Spatial Awareness in {MLLMs} via Task-Oriented Visual Supervision},
  author    = {Tang, Jiayu and Zhou, Yuchen and Gou, Chao},
  booktitle = {Proceedings of the International Joint Conference on Artificial Intelligence (IJCAI)},
  year      = {2026}
}

@article{yuan2026visionopd,
  title   = {{Vision-OPD}: Learning to See Fine Details for Multimodal {LLMs} via On-Policy Self-Distillation},
  author  = {Yuan, Qianhao and Lou, Jie and Yu, Xing and Lin, Hongyu and Sun, Le and Han, Xianpei and Lu, Yaojie},
  journal = {arXiv preprint arXiv:2605.18740},
  year    = {2026}
}

@article{li2026omniviewspace,
  title   = {{OmniView-Space}: Reinforcing Spatial Reasoning via Multi-Perspective Spatial Mapping},
  author  = {Li, Xudong and Zhang, Mengdan and Chen, Peixian and Tan, Jiaxi and Huang, Zihao and Zheng, Jingyuan and Zhang, Yan and Zheng, Xiawu and Sun, Xing and Ji, Rongrong},
  journal = {arXiv preprint arXiv:2607.00881},
  year    = {2026}
}

@article{zhou2026spatialcli,
  title   = {{SpatialCLI}: Learning to Reason With Spatial Tools, Then Without Them},
  author  = {Zhou, Yang and Huang, Zixuan and Li, Sunzhu and Yang, Zhuo and Zhang, Chen and Chen, Shunian and Yan, Caijun and Xu, Jianyao and Liu, Shunyu and Fu, Weijie and Li, Peiliang and Chen, Xiaozhi and Cai, Yuxiang},
  journal = {arXiv preprint arXiv:2607.27703},
  year    = {2026}
}

@article{xu2025mose,
  title   = {{MoSE}: Skill-by-Skill Mixture-of-Experts Learning for Embodied Autonomous Machines},
  author  = {Xu, Lu and Yu, Jiaqian and Peng, Xiongfeng and Chen, Yiwei and Li, Weiming and Yoo, Jaewook and Chunag, Sunghyun and Lee, Dongwook and Ji, Daehyun and Zhang, Chao},
  journal = {arXiv preprint arXiv:2507.07818},
  year    = {2025}
}

@article{ma2026mopd,
  title   = {{MOPD}: Multi-Teacher On-Policy Distillation for Capability Integration in {LLM} Post-Training},
  author  = {Ma, Wenhan and Wei, Jianyu and Zhao, Liang and Zhang, Hailin and Xiao, Bangjun and Li, Lei and Yang, Qibin and Gao, Bofei and Wang, Yudong and Li, Rang and Dong, Jinhao and Sui, Zhifang and Luo, Fuli},
  journal = {arXiv preprint arXiv:2606.30406},
  year    = {2026}
}

@article{wang2026vigos,
  title   = {Seeing Before Reasoning: Decoupling Perception and Reasoning for Shortcut-Resilient Multimodal On-Policy Self-Distillation},
  author  = {Wang, Sihan and Liu, Xiyao and Liu, Lianqing and Han, Zhi},
  journal = {arXiv preprint arXiv:2606.19120},
  year    = {2026}
}

@article{yin2026hopd,
  title   = {{H-OPD}: Confidence Aware Heterogeneous Multi-Teacher Multimodal On-policy Distillation},
  author  = {Yin, Qixiang and Yao, Huanjin and Cai, Yuchen and Chen, Jianghao and Wang, Ziyi and Yang, Min and Su, Fei and Zhao, Zhicheng},
  journal = {arXiv preprint arXiv:2607.02592},
  year    = {2026}
}

@inproceedings{grauman2022ego4d,
  title={{Ego4D}: Around the world in 3,000 hours of egocentric video},
  author={Grauman, Kristen and Westbury, Andrew and Byrne, Eugene and Chavis, Zachary and Furnari, Antonino and Girdhar, Rohit and Hamburger, Jackson and Jiang, Hao and Liu, Miao and Liu, Xingyu and others},
  booktitle={Proceedings of the IEEE/CVF conference on computer vision and pattern recognition (CVPR)},
  year={2022}
}

@article{wang2025internvl35,
  title={{InternVL3.5}: Advancing open-source multimodal models in versatility, reasoning, and efficiency},
  author={Wang, Weiyun and Gao, Zhangwei and Gu, Lixin and Pu, Hengjun and Cui, Long and Wei, Xingguang and Liu, Zhaoyang and Jing, Linglin and Ye, Shenglong and Shao, Jie and others},
  journal={arXiv preprint arXiv:2508.18265},
  year={2025}
}

@inproceedings{pan2023adt,
  title     = {{Aria Digital Twin}: A New Benchmark Dataset for Egocentric {3D} Machine Perception},
  author    = {Pan, Xiaqing and Charron, Nicholas and Yang, Yongqian and Peters, Scott and Whelan, Thomas and Kong, Chen and Parkhi, Omkar and Newcombe, Richard and Ren, Yuheng (Carl)},
  booktitle = {Proceedings of the IEEE/CVF International Conference on Computer Vision (ICCV)},
  year      = {2023}
}

@inproceedings{grauman2024egoexo4d,
  title     = {{Ego-Exo4D}: Understanding Skilled Human Activity from First- and Third-Person Perspectives},
  author    = {Grauman, Kristen and Westbury, Andrew and Torresani, Lorenzo and Kitani, Kris and Malik, Jitendra and Afouras, Triantafyllos and Ashutosh, Kumar and Baiyya, Vijay and Bansal, Siddhant and Boote, Bikram and others},
  booktitle = {Proceedings of the IEEE/CVF Conference on Computer Vision and Pattern Recognition (CVPR)},
  year      = {2024}
}

@inproceedings{zhang2025spar,
  title     = {From Flatland to Space: Teaching Vision-Language Models to Perceive and Reason in {3D}},
  author    = {Zhang, Jiahui and Chen, Yurui and Zhou, Yanpeng and Xu, Yueming and Huang, Ze and Mei, Jilin and Chen, Junhui and Yuan, Yu-Jie and Cai, Xinyue and Huang, Guowei and Quan, Xingyue and Xu, Hang and Zhang, Li},
  booktitle = {Advances in Neural Information Processing Systems (NeurIPS)},
  year      = {2025}
}

@misc{qwen2026qwen35,
  title  = {{Qwen3.5}: Towards Native Multimodal Agents},
  author = {{Qwen Team}},
  year   = {2026},
  url    = {https://qwen.ai/blog?id=qwen3.5}
}

@inproceedings{luo2026pyspatial,
  title     = {{pySpatial}: Generating {3D} Visual Programs for Zero-Shot Spatial Reasoning},
  author    = {Luo, Zhanpeng and Zhang, Ce and Yong, Silong and Dai, Cunxi and Wang, Qianwei and Ran, Haoxi and Shi, Guanya and Sycara, Katia and Xie, Yaqi},
  booktitle = {Proceedings of the International Conference on Learning Representations (ICLR)},
  year      = {2026}
}

@article{dang2026rynnbrain,
  title={{RynnBrain}: Open embodied foundation models},
  author={Dang, Ronghao and Guo, Jiayan and Hou, Bohan and Leng, Sicong and Li, Kehan and Li, Xin and Liu, Jiangpin and Mao, Yunxuan and Wang, Zhikai and Yuan, Yuqian and others},
  journal={arXiv preprint arXiv:2602.14979},
  year={2026}
}

@article{comanici2025gemini,
  title={Gemini 2.5: Pushing the frontier with advanced reasoning, multimodality, long context, and next generation agentic capabilities},
  author={Comanici, Gheorghe and Bieber, Eric and Schaekermann, Mike and Pasupat, Ice and Sachdeva, Noveen and Dhillon, Inderjit and Blistein, Marcel and Ram, Ori and Zhang, Dan and Rosen, Evan and others},
  journal={arXiv preprint arXiv:2507.06261},
  year={2025}
}

@misc{pichai2025gemini3,
  title        = {A New Era of Intelligence with {Gemini 3}},
  author       = {Pichai, Sundar and Hassabis, Demis and Kavukcuoglu, Koray},
  howpublished = {Google Blog},
  year         = {2025},
  url          = {https://blog.google/products-and-platforms/products/gemini/gemini-3/}
}

@techreport{openai2026gpt55,
  title       = {{GPT-5.5} System Card},
  author      = {{OpenAI}},
  institution = {OpenAI},
  year        = {2026},
  url         = {https://deploymentsafety.openai.com/gpt-5-5/gpt-5-5.pdf}
}

@inproceedings{kwon2023vllm,
  title     = {Efficient Memory Management for Large Language Model Serving with {PagedAttention}},
  author    = {Kwon, Woosuk and Li, Zhuohan and Zhuang, Siyuan and Sheng, Ying and Zheng, Lianmin and Yu, Cody Hao and Gonzalez, Joseph E. and Zhang, Hao and Stoica, Ion},
  booktitle = {Proceedings of the 29th ACM Symposium on Operating Systems Principles (SOSP)},
  year      = {2023}
}

@inproceedings{sukhbaatar2024btx,
  title   = {{Branch-Train-MiX}: Mixing Expert {LLMs} into a Mixture-of-Experts {LLM}},
  author  = {Sukhbaatar, Sainbayar and Golovneva, Olga and Sharma, Vasu and Xu, Hu and Lin, Xi Victoria and Rozi{\`e}re, Baptiste and Kahn, Jacob and Li, Daniel and Yih, Wen-tau and Weston, Jason and Li, Xian},
  booktitle = {Proceedings of the Conference on Language Modeling (COLM)},
  year    = {2024}
}

@inproceedings{shi2025flexolmo,
  title   = {{FlexOLMo}: Open Language Models for Flexible Data Use},
  author  = {Shi, Weijia and Bhagia, Akshita and Farhat, Kevin and Muennighoff, Niklas and Morrison, Jacob and Walsh, Pete and Schwenk, Dustin and Longpre, Shayne and Poznanski, Jake and Ettinger, Allyson and Liu, Daogao and Li, Margaret and Groeneveld, Dirk and Lewis, Mike and Yih, Wen-tau and Soldaini, Luca and Lo, Kyle and Smith, Noah A. and Zettlemoyer, Luke and Koh, Pang Wei and Hajishirzi, Hannaneh and Farhadi, Ali and Min, Sewon},
  booktitle = {Advances in Neural Information Processing Systems (NeurIPS)},
  year    = {2025}
}

@article{morrison2026bar,
  title   = {Train Separately, Merge Together: Modular Post-Training with Mixture-of-Experts},
  author  = {Morrison, Jacob and Adhikesaven, Sanjay and Bhagia, Akshita and Zaharia, Matei and Smith, Noah A. and Min, Sewon},
  journal = {arXiv preprint arXiv:2604.18473},
  year    = {2026}
}

@inproceedings{chen2024spatialvlm,
  title   = {{SpatialVLM}: Endowing Vision-Language Models with Spatial Reasoning Capabilities},
  author  = {Chen, Boyuan and Xu, Zhuo and Kirmani, Sean and Ichter, Brian and Sadigh, Dorsa and Guibas, Leonidas and Xia, Fei},
  booktitle = {Proceedings of the IEEE/CVF Conference on Computer Vision and Pattern Recognition (CVPR)},
  year    = {2024}
}

@inproceedings{cai2024spatialbot,
  title   = {{SpatialBot}: Precise Spatial Understanding with Vision Language Models},
  author  = {Cai, Wenxiao and Ponomarenko, Iaroslav and Yuan, Jianhao and Li, Xiaoqi and Yang, Wankou and Dong, Hao and Zhao, Bo},
  booktitle = {Proceedings of the IEEE International Conference on Robotics and Automation (ICRA)},
  year    = {2025}
}

@article{fedus2022switch,
  title   = {Switch Transformers: Scaling to Trillion Parameter Models with Simple and Efficient Sparsity},
  author  = {Fedus, William and Zoph, Barret and Shazeer, Noam},
  journal = {Journal of Machine Learning Research (JMLR)},
  year    = {2022}
}

@article{jiang2024mixtral,
  title   = {Mixtral of Experts},
  author  = {Jiang, Albert Q. and Sablayrolles, Alexandre and Roux, Antoine and Mensch, Arthur and Savary, Blanche and Bamford, Chris and Chaplot, Devendra Singh and de las Casas, Diego and Bou Hanna, Emma and Bressand, Florian and Lengyel, Gianna and Bour, Guillaume and Lample, Guillaume and Renard Lavaud, L{\'e}lio and Saulnier, Lucile and Lachaux, Marie-Anne and Stock, Pierre and Subramanian, Sandeep and Yang, Sophia and Antoniak, Szymon and Le Scao, Teven and Gervet, Th{\'e}ophile and Lavril, Thibaut and Wang, Thomas and Lacroix, Timoth{\'e}e and El Sayed, William},
  journal = {arXiv preprint arXiv:2401.04088},
  year    = {2024}
}

@inproceedings{dai2024deepseekmoe,
  title   = {{DeepSeekMoE}: Towards Ultimate Expert Specialization in Mixture-of-Experts Language Models},
  author  = {Dai, Damai and Deng, Chengqi and Zhao, Chenggang and Xu, R. X. and Gao, Huazuo and Chen, Deli and Li, Jiashi and Zeng, Wangding and Yu, Xingkai and Wu, Y. and Xie, Zhenda and Li, Y. K. and Huang, Panpan and Luo, Fuli and Ruan, Chong and Sui, Zhifang and Liang, Wenfeng},
  booktitle = {Proceedings of the Annual Meeting of the Association for Computational Linguistics (ACL)},
  year    = {2024}
}

@inproceedings{muennighoff2024olmoe,
  title   = {{OLMoE}: Open Mixture-of-Experts Language Models},
  author  = {Muennighoff, Niklas and Soldaini, Luca and Groeneveld, Dirk and Lo, Kyle and Morrison, Jacob and Min, Sewon and Shi, Weijia and Walsh, Pete and Tafjord, Oyvind and Lambert, Nathan and Gu, Yuling and Arora, Shane and Bhagia, Akshita and Schwenk, Dustin and Wadden, David and Wettig, Alexander and Hui, Binyuan and Dettmers, Tim and Kiela, Douwe and Farhadi, Ali and Smith, Noah A. and Koh, Pang Wei and Singh, Amanpreet and Hajishirzi, Hannaneh},
  booktitle = {Proceedings of the International Conference on Learning Representations (ICLR)},
  year    = {2025}
}

@inproceedings{gu2024minillm,
  title     = {{MiniLLM}: Knowledge Distillation of Large Language Models},
  author    = {Gu, Yuxian and Dong, Li and Wei, Furu and Huang, Minlie},
  booktitle = {Proceedings of the International Conference on Learning Representations (ICLR)},
  year      = {2024}
}

@inproceedings{ko2024distillm,
  title     = {{DistiLLM}: Towards Streamlined Distillation for Large Language Models},
  author    = {Ko, Jongwoo and Kim, Sungnyun and Chen, Tianyi and Yun, Se-Young},
  booktitle = {Proceedings of the International Conference on Machine Learning (ICML)},
  year      = {2024}
}

@inproceedings{ko2025distillm2,
  title     = {{DistiLLM-2}: A Contrastive Approach Boosts the Distillation of {LLMs}},
  author    = {Ko, Jongwoo and Chen, Tianyi and Kim, Sungnyun and Ding, Tianyu and Liang, Luming and Zharkov, Ilya and Yun, Se-Young},
  booktitle = {Proceedings of the International Conference on Machine Learning (ICML)},
  year      = {2025}
}

@article{wang2026madopd,
  title   = {{MAD-OPD}: Breaking the Ceiling in On-Policy Distillation via Multi-Agent Debate},
  author  = {Wang, Jianze and Liu, Ying and Chen, Jinlong and Hu, Xuchun and Zhang, Qilong and Cao, Yu and Wang, Jun and Yang, Hua and Xie, Yong and Chen, Qianglong},
  journal = {arXiv preprint arXiv:2605.01347},
  year    = {2026}
}

@article{xu2026tip,
  title   = {{TIP}: Token Importance in On-Policy Distillation},
  author  = {Xu, Yuanda and Sang, Hejian and Zhou, Zhengze and He, Ran and Wang, Zhipeng and Geramifard, Alborz},
  journal = {arXiv preprint arXiv:2604.14084},
  year    = {2026}
}

@inproceedings{cheng2025navila,
  title     = {{NaVILA}: Legged Robot Vision-Language-Action Model for Navigation},
  author    = {Cheng, An-Chieh and Ji, Yandong and Yang, Zhaojing and Gongye, Zaitian and Zou, Xueyan and Kautz, Jan and Biyik, Erdem and Yin, Hongxu and Liu, Sifei and Wang, Xiaolong},
  booktitle = {Proceedings of Robotics: Science and Systems (RSS)},
  year      = {2025}
}

@inproceedings{driess2023palme,
  title     = {{PaLM-E}: An Embodied Multimodal Language Model},
  author    = {Driess, Danny and Xia, Fei and Sajjadi, Mehdi S. M. and Lynch, Corey and Chowdhery, Aakanksha and Ichter, Brian and Wahid, Ayzaan and Tompson, Jonathan and Vuong, Quan and Yu, Tianhe and Huang, Wenlong and Chebotar, Yevgen and Sermanet, Pierre and Duckworth, Daniel and Levine, Sergey and Vanhoucke, Vincent and Hausman, Karol and Toussaint, Marc and Greff, Klaus and Zeng, Andy and Mordatch, Igor and Florence, Pete},
  booktitle = {Proceedings of the International Conference on Machine Learning (ICML)},
  year      = {2023}
}

@inproceedings{majumdar2024openeqa,
  title     = {{OpenEQA}: Embodied Question Answering in the Era of Foundation Models},
  author    = {Majumdar, Arjun and Ajay, Anurag and Zhang, Xiaohan and Putta, Pranav and Yenamandra, Sriram and Henaff, Mikael and Silwal, Sneha and Mcvay, Paul and Maksymets, Oleksandr and Arnaud, Sergio and Yadav, Karmesh and Li, Qiyang and Newman, Ben and Sharma, Mohit and Berges, Vincent and Zhang, Shiqi and Agrawal, Pulkit and Bisk, Yonatan and Batra, Dhruv and Kalakrishnan, Mrinal and Meier, Franziska and Paxton, Chris and Sax, Sasha and Rajeswaran, Aravind},
  booktitle = {Proceedings of the IEEE/CVF Conference on Computer Vision and Pattern Recognition (CVPR)},
  year      = {2024}
}

@inproceedings{yuan2025robopoint,
  title     = {{RoboPoint}: A Vision-Language Model for Spatial Affordance Prediction in Robotics},
  author    = {Yuan, Wentao and Duan, Jiafei and Blukis, Valts and Pumacay, Wilbert and Krishna, Ranjay and Murali, Adithyavairavan and Mousavian, Arsalan and Fox, Dieter},
  booktitle = {Proceedings of the Conference on Robot Learning (CoRL)},
  year      = {2025}
}

@article{xu2026deepseek,
  title={{DeepSeek-V4}: Towards Highly Efficient Million-Token Context Intelligence},
  author={{DeepSeek-AI} and Xu, Anyi and Lin, Bangcai and Xue, Bing and Wang, Bingxuan and Xu, Bingzheng and Wu, Bochao and Zhang, Bowei and Lin, Chaofan and Dong, Chen and Ling, Chenchen and others},
  journal={arXiv preprint arXiv:2606.19348},
  year={2026}
}

@inproceedings{xiao2025egoblind,
  title     = {{EgoBlind}: Towards Egocentric Visual Assistance for the Blind},
  author    = {Xiao, Junbin and Huang, Nanxin and Qiu, Hao and Tao, Zhulin and
               Yang, Xun and Hong, Richang and Wang, Meng and Yao, Angela},
  booktitle = {Advances in Neural Information Processing Systems (NeurIPS)},
  year      = {2025}
}

@article{huang2025vinci,
  title   = {{Vinci}: A Real-time Smart Assistant Based on Egocentric Vision-language Model for Portable Devices},
  author  = {Huang, Yifei and Xu, Jilan and Pei, Baoqi and Yang, Lijin and Zhang, Mingfang and He, Yuping and Chen, Guo and Chen, Xinyuan and Wang, Yaohui and Nie, Zheng and Liu, Jinyao and Lin, Dechen and Fang, Fang and Li, Kunpeng and Yuan, Chang and Qiao, Yu and Wang, Yali and Wang, Limin},
  journal = {Proceedings of the ACM on Interactive, Mobile, Wearable and Ubiquitous Technologies (IMWUT)},
  year={2025}
}

@article{liu2026objectfinder,
  title   = {{ObjectFinder}: An Open-Vocabulary Assistive System for Interactive Object Search by People Who Are Blind},
  author  = {Liu, Ruiping and Zhang, Jiaming and Sch{\"o}n, Angela and M{\"u}ller, Karin and Zheng, Junwei and Yang, Kailun and Guo, Anhong and Gerling, Kathrin and Stiefelhagen, Rainer},
  journal = {International Journal of Human-Computer Interaction (IJHCI)},
  year = {2026}
}

@inproceedings{kulkarni2026egovita,
  title     = {{EgoVITA}: Learning to Plan and Verify for Egocentric Video Reasoning},
  author    = {Kulkarni, Yogesh and Fazli, Pooyan},
  booktitle = {Proceedings of the European Conference on Computer Vision (ECCV)},
  year      = {2026}
}

@inproceedings{li2026frameoracle,
  title     = {{FrameOracle}: Learning What to See and How Much to See in Videos},
  author    = {Li, Chaoyu and Li, Tianzhi and Tao, Fei and Zhao, Zhenyu and Wu, Ziqian and Zhao, Maozheng and Song, Juntong and Niu, Cheng and Fazli, Pooyan},
  booktitle = {Proceedings of the International Conference on Machine Learning (ICML)},
  year      = {2026}
}

\clearpage
\beginappendix
\section{Training Data Generation}
\label{app:data_construction}

Section~\ref{sec:experimental_setup} summarizes the data sources, generation models, and sample counts.
Here, we describe how annotations and generated trajectories form role supervision, followed by the detailed data composition.

\subsection{Data Construction}
\label{app:data_sources}

ADT~\citep{pan2023adt} provides camera and object poses together with geometry and identity annotations, from which we derive answers to spatial subquestions about distance, direction, size, location, and identity.
Ego-Exo4D~\citep{grauman2024egoexo4d}, VSI-590K~\citep{yang2025cambrians}, and SPAR-7M~\citep{zhang2025spar} supply original questions and final answers, including multiple-choice questions.
Ego-Exo4D also provides annotated subjective questions, and Ego4D~\citep{grauman2022ego4d} contributes additional Planner samples.
Annotations for the original question support trajectory verification but do not independently verify every intermediate observation.

We construct demonstrations for both \texttt{DIRECT} and \texttt{PLAN}.
For \texttt{DIRECT}, the frozen teacher backbone answers the original video and question with no LoRA active.
For \texttt{PLAN}, the planning model generates subquestions and receives observations before deciding whether to request further evidence or compose the final answer.
Available subquestion annotations are inserted as answers to the corresponding subquestions.
When an observation must be generated, the visual model receives only the video and subquestion, without the original answer options or other planning history.
Generated observations without independent annotations serve as weak supervision.

We compare each generated final answer with the task annotation and allow up to three retries for an incorrect \texttt{PLAN} trajectory.
If the resulting \texttt{PLAN} answer is correct, we retain it as a \texttt{PLAN} example regardless of the frozen base's answer.
If \texttt{PLAN} remains incorrect but the frozen base answers correctly, we retain a \texttt{DIRECT} example.
We discard examples for which both routes are incorrect.
From retained \texttt{PLAN} trajectories, we extract subquestion-answer pairs for Perception and planning or composition targets with their preceding histories for the Planner.
Appendix~\ref{app:prompts} describes the generation prompts and their conversion into training conversations.

\subsection{Data Composition}
\label{app:data_composition}

\begin{table}[h]
\centering
\small
\caption{Source composition of the teacher training sets. All values count samples, and dashes indicate no samples from that source.}
\label{tab:teacher_data_sources}
\begin{tabular}{lrrrr}
\toprule
& \multicolumn{2}{c}{SFT} & \multicolumn{2}{c}{GRPO} \\
\cmidrule(lr){2-3}\cmidrule(lr){4-5}
Source & Perception & Planner & Perception & Planner \\
\midrule
ADT & 27,279 & 6,926 & 1,085 & 1,900 \\
Ego-Exo4D & 1,167 & 2,600 & 1,004 & 778 \\
VSI-590K & 5,502 & 3,777 & 535 & 896 \\
SPAR-7M & 1,863 & 1,792 & 536 & 298 \\
Ego4D & -- & 2,722 & -- & -- \\
\midrule
Total & 35,811 & 17,817 & 3,160 & 3,872 \\
\bottomrule
\end{tabular}
\end{table}

\begin{table}[h]
\centering
\small
\caption{Question types in the Perception SFT set. Counts sum to 35,811 samples.}
\label{tab:perception_sft_facets}
\begin{tabular}{lrlr}
\toprule
Type & Samples & Type & Samples \\
\midrule
Bearing & 12,283 & Count & 1,330 \\
Distance & 11,339 & Other & 1,820 \\
Camera distance & 2,732 & Identity & 711 \\
Size & 2,609 & Egomotion bearing & 580 \\
Object location & 2,255 & Egomotion distance & 152 \\
\bottomrule
\end{tabular}
\end{table}

Table~\ref{tab:teacher_data_sources} reports the source composition of the teacher training sets, and Table~\ref{tab:perception_sft_facets} summarizes the Perception SFT question types.
Perception GRPO includes 2,089 samples with objective subquestion annotations from ADT and Ego-Exo4D, and 1,071 without such annotations from SPAR-7M and VSI-590K.
The latter have multiple-choice original questions with
annotated correct options, which support the observation utility rewards defined in Appendix~\ref{app:teacher_training}.
Planner GRPO also uses multiple-choice questions with annotated correct options.

Student initialization samples are selected through correctness filtering.
The Planner subset comprises 2,566 planning samples and 1,534 composition samples.
These samples cover both routes, with initial choices included among the planning targets.
Joint SFT combines complete interaction samples from both routes with reused subquestion samples.
The MOPD mixture contains 6,500 spatial questions and 1,500 general (non-spatial) questions.
The latter are randomly sampled from the teacher training sets and cover tasks such as action recognition.
A source question can yield several subquestion samples and can be reused across stages, so the reported counts do not represent disjoint collections of questions.
We verified that our training data contain none of the source recordings used for benchmark evaluation.

\section{Additional Implementation Details}
\label{app:implementation}

This section specifies the teacher rewards, student parameterization and training objectives, and inference procedure.

\subsection{Teacher Training}
\label{app:training_details}
\label{app:teacher_training}

\paragraph{Training order.}
Section~\ref{sec:experimental_setup} specifies the teacher backbone and adapter configuration.
Both roles first undergo SFT.
We supervise the Planner's initial \texttt{DIRECT}/\texttt{PLAN} choice and, for \texttt{PLAN}, its subquestions and final composition.
The frozen base's direct answer is excluded from Planner supervision.
During Perception GRPO, the current Perception policy samples a group of answers to each subquestion.
Objective annotations provide verifier rewards where available.
For answers without a ground-truth subquestion answer, the frozen SFT-trained Planner, denoted by $\bar\pi_{\mathrm{PC}}$, provides utility scores during training as detailed below.
Only the Perception adapter is updated in this stage.
After training Perception, we freeze the resulting policy and optimize the Planner through the complete interaction in Equation~\ref{eq:interaction}.
Each rollout begins with the current Planner's mode choice.
A \texttt{DIRECT} rollout obtains its answer from the frozen base.
A \texttt{PLAN} rollout continues with Planner subquestions, frozen Perception observations, and Planner composition.
A sampled group for one original question can contain both modes.
Each role uses its own SFT policy as the GRPO KL reference.
For \texttt{DIRECT}, the Planner loss and KL masks cover only its initial choice.
For \texttt{PLAN}, they also cover its subquestions and composition.
Frozen base and Perception outputs are excluded.
The final frozen policies $\pi_{\mathrm{PC}}$ and $\pi_{\mathrm V}$ subsequently supervise student training.

\paragraph{Objective perception rewards.}
For an observation $o$ with an independent subquestion annotation $z$, the reward is
\begin{equation}
\label{eq:objective_reward}
 R_{\mathrm{obj}}(o,z)
 =\mathbf 1[\operatorname{valid}(o)]\,V_{\kappa}(o,z),
\end{equation}
where validity requires parsing as the requested answer type $\kappa$.
All verifier scores lie in $[0,1]$.
Size questions are included in SFT but not in objective Perception GRPO.
We use the following rules:
\begin{itemize}
    \item \textbf{Bearing.} Index the eight directions around a circle.
    For predicted and annotated indices $i_o,i_z$, let $d{=}\min(|i_o-i_z|,8-|i_o-i_z|)$.
    The scores for $d{=}0,1,2$, and $d\geq3$ are $1.00$, $0.60$, $0.18$, and $0$, respectively.
    \item \textbf{Distance.} For a nonzero annotated distance, define $e{=}|o-z|/|z|$.
    The mean relative accuracy is
    $V_{\mathrm{dist}}(o,z){=}\frac{1}{11}\sum_{k=0}^{10}\mathbf 1[e\leq1-c_k]$,
    where $c_k{=}0.50+0.045k$.
    \item \textbf{Identity.} Compute precision and recall from normalized token overlap and take their F1 score.
    To discourage listing multiple candidate identities, apply an additional length penalty by multiplying this score by $|z|/(|z|+\mathrm{extra})$, where $\mathrm{extra}=\max(0,|o|-|z|)$ and $|o|,|z|$ count normalized predicted and reference tokens, respectively.
    \item \textbf{Count and presence.} Counts receive $\exp(-|o-z|)$.
    Presence receives $\mathbf 1[o{=}z]$ after parsing.
    \item \textbf{Location.} Average the bearing and distance scores for the corresponding components.
\end{itemize}

\paragraph{Answer margins.}
Both the observation utility and Planner reward use the margin of the correct final option.
For answer options $\mathcal A$, annotated option $a^\star$, policy $\pi$, and context $C$, define
\begin{equation}
\label{eq:answer_margin}
 M_{\pi}(C)
 =\ell_{\pi}(a^\star\mid C)
 -\log\sum_{a\in\mathcal A\setminus\{a^\star\}}
 \exp\bigl(\ell_{\pi}(a\mid C)\bigr),
\end{equation}
where $\ell_{\pi}(a\mid C)$ is the log probability of the option label.
The options in $\mathcal A$ are the candidate answers to the original question.
The correct-option annotation is used for scoring and is withheld from generation prompts.
Perception answers the intermediate subquestions with numbers or short text.
At inference, the model can also answer questions without predefined options.
Throughout this appendix, $\operatorname{clip}(z,a,b){=}\min\{b,\max\{a,z\}\}$.

\paragraph{Observation utilities without ground-truth subquestion answer.}
The original question has an annotated final answer even when its subquestion has no ground-truth.
We use this final answer to evaluate each answer sampled by the current Perception policy.
Perception receives only the video and subquestion.
It does not receive the original question's answer options or annotated answer.

For a sampled answer $o$, let $C[o]$ contain the video, original question and options, and the observations used for composition, with $o$ filling the slot for the current subquestion.
The comparison context $C[\texttt{unknown}]$ replaces only that answer with \texttt{unknown}.
We force the frozen SFT-trained Planner to compose on these two contexts and score the final answer options using Equation~\ref{eq:answer_margin}.
The video and all other evidence are identical, and the Planner neither selects a mode nor asks further subquestions during this evaluation.
The utility of the sampled answer is
\begin{equation}
\label{eq:slot_utility}
 \begin{aligned}
 \delta(o)
 &=M_{\bar\pi_{\mathrm{PC}}}(C[o])
 -M_{\bar\pi_{\mathrm{PC}}}(C[\texttt{unknown}]),\\
 u_{+}(o)
 &=\operatorname{clip}\!\left(\frac{\delta(o)-\epsilon}{s_{\mathrm V}},0,1\right),\\
 u_{-}(o)
 &=\operatorname{clip}\!\left(\frac{-\delta(o)-\epsilon}{s_{\mathrm V}},0,1\right).
 \end{aligned}
\end{equation}
Positive $\delta(o)$ means that providing the answer increases the correct option's margin, while negative $\delta(o)$ means that it reduces the margin.
The dead zone $\epsilon$ excludes small changes, and $s_{\mathrm V}$ sets their scale.
Every newly sampled answer is scored directly during training.
The \texttt{unknown} margin is common to answers in the same group and can be reused while the full input context and frozen Planner remain unchanged.
The subjective reward is
\begin{equation}
\label{eq:subjective_reward}
 R_{\mathrm{subj}}(o)
 =\gamma n
 \begin{cases}
 \alpha_{+}u_{+}(o)-\alpha_{-}u_{-}(o)
 -\beta_{\mathrm{leak}}^{\mathrm V}I_{\mathrm{leak}}(o),
 &\operatorname{valid}(o),\\
 -\beta_{\mathrm{invalid}}^{\mathrm V},&\text{otherwise}.
 \end{cases}
\end{equation}
For each initially correct prompt, we remove the reference observation and recompose the final answer with the frozen Planner.
We set the offline necessity weight $n_{\mathrm{pre}}$ to $1.0$ if this removal changes the correct answer and to $0.1$ otherwise.
When no offline necessity value is available, the data builder uses a default value of $0.25$.
For subquestions supervised by teacher-generated references rather than ground-truth answers, we multiply this weight by $0.5$ before GRPO.
The resulting value is stored with each training sample and used as $n$ in Equation~\ref{eq:subjective_reward}.
Thus, $n$ is $0.5$, $0.05$, or $0.125$ for necessary, redundant, or default teacher-referenced samples, respectively.
If a training sample lacks a stored necessity weight, the reward computation uses $n=0.25$ as a fallback.
$I_{\mathrm{leak}}$ detects explicit selection of a multiple-choice option label.
The outer factor $\gamma n$ also scales both penalties.
Rewards within each sampled group are used to compute GRPO advantages and update the Perception adapter.
Subsequent groups are sampled from the updated Perception policy, while the Planner remains frozen throughout this stage.
This reward provides weak supervision for an observation's usefulness during composition under fixed evidence.
GRPO regularizes the Perception policy toward its SFT checkpoint, and the weak reward penalizes invalid outputs and explicit option-label answers.
Objective verifiers provide supervision for subquestion accuracy where independent annotations are available.

\paragraph{Planner reward.}
We score the complete path selected by the current Planner's initial action $b$.
Let $r_{\mathrm{out}}{=}\mathbf 1[\hat a{=}a^\star]$ denote final correctness.
The direct margin $M_{\mathrm{direct}}$ uses the frozen teacher backbone $\pi_{\mathrm B}$ with no LoRA active on the original video and question, including any answer options, with no intermediate observations or Planner-private text.
For $b{=}\texttt{PLAN}$, the selected-path margin $M_{\mathrm{route}}$ uses the current Planner's composition with observations from the frozen Perception teacher.
For $b{=}\texttt{DIRECT}$, it uses the same frozen base and identical context as $M_{\mathrm{direct}}$, so $M_{\mathrm{route}}=M_{\mathrm{direct}}$.
For $d{=}(M_{\mathrm{route}}-M_{\mathrm{direct}})/s_{\mathrm{PC}}$, the reward is
\begin{equation}
\label{eq:planner_reward}
 \begin{aligned}
 R_{\mathrm{PC}}
 &=r_{\mathrm{out}}
 +\lambda_{\mathrm{gain}}(1-r_{\mathrm{out}})\operatorname{clip}(d,0,1)\\
 &\quad-\lambda_{\mathrm{reg}}\operatorname{clip}(-d,0,1)
 -P_{\mathrm{protocol}}.
 \end{aligned}
\end{equation}
Correctness supplies the principal signal.
Positive margin gain provides additional feedback when the final answer is incorrect.
Regression below the direct base policy is penalized regardless of correctness.
For \texttt{PLAN}, the gain includes both specialization and evidence collection because the direct reference is a different policy.
For \texttt{DIRECT}, the margin difference is zero.
The protocol penalty is
\begin{equation}
\label{eq:protocol_penalty}
 P_{\mathrm{protocol}}
 =\beta_{\mathrm{parse}}^{\mathrm{PC}}I_{\mathrm{parse}}
 +\beta_{\mathrm{ref}}^{\mathrm{PC}}I_{\mathrm{ref}}
 +\beta_{\mathrm{leak}}^{\mathrm{PC}}I_{\mathrm{leak}}.
\end{equation}
The indicators identify an invalid mode or interaction structure, unresolved references in a subquestion, and disclosure of answer options through the perception interface.
Validity follows the selected mode's protocol: a \texttt{DIRECT} path is valid without subquestions, and the reference and interface penalties apply only when those components occur.
These penalties apply regardless of the final answer.
Table~\ref{tab:training_coefficients} summarizes reward and training coefficients.

\begin{table}[t]
\centering
\caption{Reward and training coefficients.
$n_{\mathrm{pre}}$ denotes the necessity weight before reference downweighting.}
\label{tab:training_coefficients}
\small
\renewcommand{\arraystretch}{1.4}
\begin{tabular}{lll}
\toprule
\textbf{Component} & \textbf{Parameter} & \textbf{Value} \\
\midrule
LoRA alpha (teachers / student residuals)
& $\alpha$ & $32$ \\
GRPO KL coefficient (Planner / Perception)
& $\beta_{\mathrm{KL}}$ & $0.04$ \\
No-ground-truth reward scale
& $\gamma$ & $0.5$ \\
Necessity (necessary / redundant / default)
& $n_{\mathrm{pre}}$ & $1.0,\ 0.1,\ 0.25$ \\
Teacher-reference downweight
& $m_{\mathrm{ref}}$ & $0.5$ \\
Missing-weight fallback
& $n_{\mathrm{fallback}}$ & $0.25$ \\
Positive / negative utility
& $\alpha_{+},\alpha_{-}$ & $0.60,\ 0.30$ \\
Perception leakage penalty
& $\beta_{\mathrm{leak}}^{\mathrm V}$ & $0.50$ \\
Perception invalid-answer penalty
& $\beta_{\mathrm{invalid}}^{\mathrm V}$ & $1$ \\
Utility dead zone
& $\epsilon$ & $0.10$ \\
Utility scale
& $s_{\mathrm V}$ & $2$ \\
Planner margin scale
& $s_{\mathrm{PC}}$ & $2$ \\
Planner gain / regression
& $\lambda_{\mathrm{gain}},\lambda_{\mathrm{reg}}$ & $0.0875,\ 0.10$ \\
Invalid interaction structure
& $\beta_{\mathrm{parse}}^{\mathrm{PC}}$ & $0.25$ \\
Unresolved reference
& $\beta_{\mathrm{ref}}^{\mathrm{PC}}$ & $0.10$ \\
Planner interface leakage
& $\beta_{\mathrm{leak}}^{\mathrm{PC}}$ & $0.30$ \\
Sequence / ground-truth supervision
& $\lambda_{\mathrm{seq}},\lambda_{\mathrm{gt}}$ & $1,\ 1$ \\
\bottomrule
\end{tabular}
\end{table}

\subsection{Student Training}
\label{app:student_training}

\paragraph{Experts and trainable parameters.}
For both student backbones in Section~\ref{sec:experimental_setup}, we replace the feedforward networks in the last 12 of the 36 decoder layers with role experts.
These are layers $24$-$35$ under zero-based indexing.
In each selected layer $\ell$, the original feedforward network $F_0^{(\ell)}$ is copied into three experts, with one active expert per token outside the common prefix:
\begin{equation}
\label{eq:role_experts}
 \operatorname{FFN}^{(\ell)}(h_t,r_t)=E_{r_t}^{(\ell)}(h_t),
 \qquad
 E_r^{(\ell)}\big|_{\mathrm{init}}=F_0^{(\ell)},
 \quad r\in\{\mathrm G,\mathrm{PC},\mathrm V\}.
\end{equation}
The segment role determines $r_t$.
In the same layers, rank-16 role attention residuals modify \texttt{q\_proj}, \texttt{k\_proj}, \texttt{v\_proj}, and \texttt{o\_proj}.
Outside the common prefix, each such projection uses $W_r^{(\ell)}{=}W_0^{(\ell)}+B_r^{(\ell)}A_r^{(\ell)}$, with the scaling absorbed into $B_r^{(\ell)}$.
The PC and V residuals use $B_r^{(\ell)}{=}0$ at initialization.
These copies and zero residuals preserve the base computation for the same input context at initialization.
We freeze the vision encoder, aligner, and original language parameters.
During initialization, the PC and V experts and their attention residuals are trainable.
The general feedforward expert remains frozen throughout training.
At the start of MOPD, we add a general attention residual for \texttt{DIRECT} answers, using the same rank, projections, and layers as PC and V.
MOPD trains this residual together with the PC and V parameters.
All role attention residuals are disabled on the common video prefix, which always uses the frozen base path.

\paragraph{Initialization supervision.}
For role $r$, let $\mathcal D_{\mathrm{seq}}^r$ contain teacher-generated targets and $\mathcal D_{\mathrm{gt}}^r$ contain available ground-truth targets.
The components of the initialization objective are
\begin{equation}
\label{eq:initialization_ce}
 \mathcal L_k^r
 =-\mathbb E_{(C,z,\mathcal I)\sim\mathcal D_k^r}
 \left[
 \frac{1}{|\mathcal I|}
 \sum_{t\in\mathcal I}
 \log\pi_\theta^r(z_t\mid C,z_{<t})
 \right],
 \quad k\in\{\mathrm{seq},\mathrm{gt}\}.
\end{equation}
The sequence targets are generated greedily by the corresponding teacher and inserted as fixed targets into the student's template.
PC sequence targets cover the initial mode choice in both routes and subquestions and composition under \texttt{PLAN}, while V sequence targets cover subquestion answers.
\texttt{DIRECT} answer tokens belong to G and are excluded from initialization supervision.
Ground-truth targets supervise annotated subquestion answers and final option labels on the corresponding V and PC segments.
The two supervision sources use separate samples and target masks, so their losses do not count the same supervised positions twice.
We set $w_{\mathrm{PC}}{=}w_{\mathrm V}{=}1$, with each role loss averaged over its supervised token positions as in Equation~\ref{eq:initialization_ce}.
The roles first update their own parameter sets separately.
We then continue joint SFT on subquestions and complete interaction samples from both modes, applying the same objective to each PC or V role's target positions.
The joint SFT sampler balances sample counts between PC and V.
G is not trained during initialization or joint SFT.
The general samples used for MOPD are part of its prompt mixture, as reported in Appendix~\ref{app:data_composition}.

\paragraph{Teacher assignment and scoring.}
During MOPD, the student samples one trajectory per prompt with temperature $1.0$ and top-$p$ $1.0$.
Each trajectory is generated in full before segment-wise backpropagation.
We accumulate gradients over all segments from 16 sampled trajectories before each optimizer update.
For a student-generated segment with role $r$, the teacher is
\begin{equation}
\label{eq:teacher_assignment}
 T(r)=
 \begin{cases}
 \pi_{\mathrm{PC}}, &r=\mathrm{PC},\\
 \pi_{\mathrm V}, &r=\mathrm V,\\
 \pi_{\mathrm B}, &r=\mathrm G.
 \end{cases}
\end{equation}
The initial mode choice, planning, and composition share the PC teacher, including supervision of the final answer in \texttt{PLAN}.
A \texttt{DIRECT} trajectory has a PC mode-choice segment followed by a G answer segment, supervised by $\pi_{\mathrm{PC}}$ and $\pi_{\mathrm B}$, respectively.
For both student backbones, $\pi_{\mathrm B}$ is the frozen SenseNova-SI teacher with both LoRA adapters disabled.
It supervises the General expert's direct answers.
Each new rollout is scored live by its assigned teachers.
For each sampled token, its teacher log probability is obtained from its logit minus the log-sum-exp over the complete vocabulary.
Scoring uses the student's sampled tokens and preceding context, not a teacher-generated continuation.
The student log probability in Equation~\ref{eq:distillation_advantage} corresponds to the student parameters $\bar\theta$ used to generate the trajectory.
The advantage is detached and clipped to $[-5,5]$, with $A_{\max}{=}5$, before weighting the current student's log probability.
We set $\omega_{\mathrm{PC}}{=}\omega_{\mathrm V}{=}\omega_{\mathrm G}{=}1$.
Equation~\ref{eq:segment_mopd} normalizes by the total supervised token count for each role across the trajectory.
Thus, a role's contribution does not grow simply because it generates more or longer segments.

\paragraph{Local reverse-KL interpretation.}
For a fixed role $r$ and context $C$, let $q_\theta(z)=\pi_\theta^r(z\mid C)$, $\tau(z)=T(r)(z\mid C)$, and $A(z)=\log[\tau(z)/q_{\bar\theta}(z)]$.
Using $\mathbb E_{z\sim q_\theta}[\nabla_\theta\log q_\theta(z)]=0$, we obtain
\begin{equation}
\label{eq:local_reverse_kl}
\left.\nabla_\theta D_{\mathrm{KL}}(q_\theta\|\tau)\right|_{\theta=\bar\theta}
=-\mathbb E_{z\sim q_{\bar\theta}}
\left[A(z)\left.\nabla_\theta\log q_\theta(z)\right|_{\theta=\bar\theta}\right].
\end{equation}
Thus, the unclipped token loss $-\operatorname{sg}[A(z)]\log q_\theta(z)$ matches this gradient in expectation at $\theta=\bar\theta$, before role-length weighting.
At $\theta\ne\bar\theta$, the fixed-snapshot update need not match the current policy's local reverse-KL gradient.
Clipping bounds extreme token weights, while the trajectory-dependent factor $1/N_r(Y)$ averages each role's contribution.
These operations modify the gradient, so Equation~\ref{eq:segment_mopd} is a local surrogate evaluated on fixed sampled contexts.

\paragraph{Context boundaries.}
Perception receives the video, the current self-contained subquestion, and the preceding tokens of its own answer.
It does not receive the original answer options or other interaction history.
The Planner receives the video, original question, and accumulated history available before each output.
For a \texttt{DIRECT} answer, the frozen base teacher and student G role receive the original video and question, including any answer options, without the Planner's private prefix or mode-choice text.
Rollout generation, teacher scoring, and student loss computation use the same information boundaries, with consistent token positions within each model's template.
For \texttt{DIRECT}, $\pi_{\mathrm B}$ uses its native answer template, while G uses the shared prefix and G-role template. The G attention residual is trained to align their output distributions.
Completed-sequence scoring can process known target tokens in parallel, while the interaction itself remains autoregressive.
Teacher adapters are activated separately for scoring, with a fresh decoder prefill for the corresponding role context.
For direct base scoring, the PC adapter is disabled and the original video--question context receives a fresh base prefill.
Shared frozen visual features are cached, but decoder KV states are not reused across teacher adapter changes.

\subsection{Inference}
\label{app:inference_details}

\paragraph{Routing and interaction.}
The initial tool name determines the mode: \texttt{answer\_with\_base} selects \texttt{DIRECT}, while \texttt{ask\_perception} selects \texttt{PLAN}.
For \texttt{PLAN}, the remainder of the initial request contains the first subquestion or subquestions.
Both modes are available on every benchmark.
The execution loop parses \texttt{answer\_with\_base} to activate G for direct answering, or \texttt{ask\_perception} to activate V for the requested observations.
Each request $u_j$ contains one or more \texttt{<tool\_call>} blocks, each closed by \texttt{</tool\_call>}.
Each block contains an \texttt{ask\_perception} call with one \texttt{sub\_question} argument.
Independent subquestions may be issued in the same request.
A subquestion that depends on an earlier observation is generated in a subsequent Planner turn, with the required information included explicitly in its wording.
After all calls in the request have completed, the subquestions and their corresponding observations are appended to the Planner's history in request order.
Execution then returns to PC, which can request further evidence or compose the final answer.
No external teacher or spatial tool is called during student inference.

\paragraph{Interaction distribution.}
We formalize the interaction protocol introduced in Section~\ref{sec:method} for both the teacher system and the student.
Let $\rho_{\mathrm{PC}}$, $\rho_{\mathrm V}$, and $\rho_{\mathrm D}$ denote the planning-and-composition, perception, and direct answering policies of the system under consideration.
For the teachers, these policies are $(\pi_{\mathrm{PC}},\pi_{\mathrm V},\pi_{\mathrm B})$.
For the student, they are $(\pi_\theta^{\mathrm{PC}},\pi_\theta^{\mathrm V}, \pi_\theta^{\mathrm G})$.

At planning round $j$, the request $u_j$ contains $m_j\geq1$ self-contained subquestions $\tilde q_{j,1},\ldots,\tilde q_{j,m_j}$.
Perception produces the corresponding answers $o_j{=}(o_{j,1},\ldots,o_{j,m_j})$.
The interaction history after $j$ rounds is $H_j{=}((u_1,o_1),\ldots,(u_j,o_j))$, with $H_0{=}()$.

Let $p(b,y\mid x)$ denote the probability of the initial mode choice and complete output given input $x$.
Let $p_{\mathrm{int}}(y\mid x,b)$ denote the conditional distribution of the interaction under \texttt{PLAN}.
These distributions factorize as
\begin{equation}
\label{eq:interaction}
\begin{aligned}
 p(b,y\mid x)
 &=\rho_{\mathrm{PC}}(b\mid x)
 \begin{cases}
 \rho_{\mathrm D}(a\mid x), & b=\texttt{DIRECT},\\
 p_{\mathrm{int}}(y\mid x,b), & b=\texttt{PLAN},
 \end{cases}\\
 p_{\mathrm{int}}(y\mid x,b)
 &=\prod_{j=1}^{J}\left[
 \rho_{\mathrm{PC}}(u_j\mid x,b,H_{j-1})
 \prod_{k=1}^{m_j}
 \rho_{\mathrm V}(o_{j,k}\mid v,\tilde q_{j,k})
 \right]\\
 &\quad\times
 \rho_{\mathrm{PC}}(c\mid x,b,H_J).
\end{aligned}
\end{equation}
Here $\rho_{\mathrm{PC}}(b\mid x)$ scores the initial tool choice: \texttt{answer\_with\_base} selects \texttt{DIRECT}, while \texttt{ask\_perception} selects \texttt{PLAN}.
For $b=\texttt{PLAN}$, $u_1$ denotes the remaining tokens of the initial request after the route-selecting tool name, so the two factors cover disjoint token positions.
Subsequent Planner outputs either request additional observations or compose the final answer, which determines $J$ during generation.
The total number of subquestions is $\sum_{j=1}^{J}m_j$.
For the student, the resulting trajectory distribution is denoted by $p_\theta$, while the snapshot used for MOPD rollouts induces $p_{\bar\theta}$.

\paragraph{Common prefix and KV reuse.}
The student encodes the video once and prefills a common prefix containing only information shared by all roles.
Prefix tokens follow the frozen base path: the general feedforward expert is used, and every role attention residual is disabled at these positions.
Consequently, the prefix computation is independent of the current role and subquestion.
Role instructions, the original question, and answer options occur after the common prefix, within their permitted contexts.
All student roles reuse the prefix KV states while processing their own text contexts.
The same prefix path, visibility rules, and position assignments are used in training and inference.
Perception therefore reuses the video prefix without gaining access to the Planner's private text context.
After a \texttt{DIRECT} choice, G reuses the frozen video prefix and processes its own original-question context with the general attention residual trained during MOPD.
It does not reuse the PC text-suffix KV states.
The teacher comparison also caches the shared visual features, but prefills the decoder context separately for its role adapters and for the base with adapters disabled.
The student completes the interaction autoregressively within one checkpoint without adapter switching.

\section{Additional Experimental Details and Results}
\label{app:additional_experiments}

This section supplements the experiments in Section~\ref{sec:experiments} with bootstrap confidence intervals, complete results by subtask, direct-answer training baselines, inference efficiency, and gradient analysis of multi-teacher distillation.

\subsection{Statistical Analysis}
\label{app:statistical_analysis}

We compare each student with its corresponding base model using a paired bootstrap over videos with 10,000 resamples.
Each resample includes all questions from the sampled videos, with identical sampling indices used for both models.
We recompute the benchmark scores for each resample and obtain the 95\% confidence interval from the 2.5th and 97.5th percentiles of the student-minus-base score differences.
As shown in Table~\ref{tab:bootstrap_ci}, all six intervals exclude zero, indicating statistically significant improvements across both backbones and all three benchmarks.

\begin{table}[t]
\centering
\small
\caption{Student improvements over their corresponding base models
and 95\% paired bootstrap confidence intervals.
Gains and intervals are in percentage points.}
\label{tab:bootstrap_ci}
\begin{tabular}{llcc}
\toprule
\textbf{Backbone} & \textbf{Benchmark} &
\textbf{Gain} & \textbf{95\% CI} \\
\midrule
Qwen3-VL-8B
& VSI-Bench & 4.2 & [+3.4, +5.0] \\
& EgoProx   & 14.9 & [+12.7, +17.1] \\
& R3D-Bench & 8.5 & [+7.2, +9.7] \\
\midrule
SenseNova-SI-1.3-8B
& VSI-Bench & 2.1 & [+1.6, +2.6] \\
& EgoProx   & 13.0 & [+10.8, +15.3] \\
& R3D-Bench & 7.4 & [+6.3, +8.5] \\
\bottomrule
\end{tabular}
\end{table}

\begin{table}[t]
\centering
\caption{\textbf{Results by subtask on VSI-Bench.}
Scores (\%) use mean relative accuracy (MRA) for numerical questions and accuracy for multiple-choice questions.
$n$ is the number of evaluated questions. \textbf{Bold} indicates student scores. $\Delta$ is the improvement over the corresponding base.}
\label{tab:subtask_vsi}
\small
\setlength{\tabcolsep}{3pt}
\renewcommand{\arraystretch}{1.1}
\begin{tabular*}{\textwidth}{@{\extracolsep{\fill}}lrrrrrrr}
\toprule
& & \multicolumn{3}{c}{\textbf{Qwen3-VL-8B}}
  & \multicolumn{3}{c}{\textbf{SenseNova-SI-1.3-8B}} \\
\cmidrule(lr){3-5}\cmidrule(lr){6-8}
\textbf{Subtask} & $n$
& Base & \methodname{} & $\Delta$
& Base & \methodname{} & $\Delta$ \\
\midrule
Object Relative Direction
& 968 & 57.23 & \textbf{58.88} & $+1.65$
& 63.33 & \textbf{63.74} & $+0.41$ \\
Object Size Estimation
& 953 & 84.66 & \textbf{85.04} & $+0.38$
& 84.93 & \textbf{85.05} & $+0.12$ \\
Object Absolute Distance
& 834 & 42.49 & \textbf{49.65} & $+7.16$
& 45.97 & \textbf{50.36} & $+4.39$ \\
Object Relative Distance
& 710 & 48.03 & \textbf{53.66} & $+5.63$
& 59.15 & \textbf{62.39} & $+3.24$ \\
Object Appearance Order
& 618 & 69.42 & \textbf{70.23} & $+0.81$
& 74.60 & \textbf{74.92} & $+0.32$ \\
Object Counting
& 565 & 64.99 & \textbf{67.38} & $+2.39$
& 79.56 & \textbf{79.88} & $+0.32$ \\
Room Size Estimation
& 288 & 61.84 & \textbf{66.60} & $+4.76$
& 67.05 & \textbf{69.38} & $+2.33$ \\
Route Planning
& 194 & 34.54 & \textbf{45.36} & $+10.82$
& 43.81 & \textbf{49.48} & $+5.67$ \\
\midrule
\textbf{Overall}
& 5,130 & 57.90 & \textbf{62.10} & $+4.20$
& 64.80 & \textbf{66.90} & $+2.10$ \\
\bottomrule
\end{tabular*}
\end{table}

\begin{table}[t]
\centering
\caption{\textbf{Results by subtask on EgoProx.} 
Scores are multiple-choice accuracy (\%).
$n$ is the number of evaluated questions. \textbf{Bold} indicates student scores. $\Delta$ is the improvement over the corresponding base. }
\label{tab:subtask_egoprox}
\small
\setlength{\tabcolsep}{3pt}
\renewcommand{\arraystretch}{1.1}
\begin{tabular*}{\textwidth}{@{\extracolsep{\fill}}lrrrrrrr}
\toprule
& & \multicolumn{3}{c}{\textbf{Qwen3-VL-8B}}
  & \multicolumn{3}{c}{\textbf{SenseNova-SI-1.3-8B}} \\
\cmidrule(lr){3-5}\cmidrule(lr){6-8}
\textbf{Subtask} & $n$
& Base & \methodname{} & $\Delta$
& Base & \methodname{} & $\Delta$ \\
\midrule
Exploitation Relative
& 693 & 34.63 & \textbf{46.75} & $+12.12$
& 34.92 & \textbf{49.64} & $+14.72$ \\
Exploitation Approx
& 422 & 37.68 & \textbf{46.21} & $+8.53$
& 41.47 & \textbf{47.16} & $+5.69$ \\
Intention Approx
& 386 & 32.64 & \textbf{42.23} & $+9.59$
& 41.45 & \textbf{49.48} & $+8.03$ \\
Intention Relative
& 342 & 29.24 & \textbf{52.34} & $+23.10$
& 30.70 & \textbf{50.58} & $+19.88$ \\
Exploration Relative
& 191 & 17.28 & \textbf{46.60} & $+29.32$
& 15.71 & \textbf{39.79} & $+24.08$ \\
Exploration Approx
& 187 & 16.58 & \textbf{36.90} & $+20.32$
& 25.67 & \textbf{34.76} & $+9.09$ \\
\midrule
\textbf{Overall}
& 2,221 & 31.02 & \textbf{45.88} & $+14.86$
& 34.22 & \textbf{47.19} & $+12.97$ \\
\bottomrule
\end{tabular*}
\end{table}

\begin{table}[t]
\centering
\caption{\textbf{Results by subtask on R3D-Bench.}
Scores (\%) use mean relative accuracy (MRA) for numerical questions and accuracy for multiple-choice questions.
$n$ is the number of evaluated questions. \textbf{Bold} indicates student scores. $\Delta$ is the improvement over the corresponding base.}
\label{tab:subtask_r3d}
\small
\setlength{\tabcolsep}{3pt}
\renewcommand{\arraystretch}{1.1}
\begin{tabular*}{\textwidth}{@{\extracolsep{\fill}}lrrrrrrr}
\toprule
& & \multicolumn{3}{c}{\textbf{Qwen3-VL-8B}}
  & \multicolumn{3}{c}{\textbf{SenseNova-SI-1.3-8B}} \\
\cmidrule(lr){3-5}\cmidrule(lr){6-8}
\textbf{Subtask} & $n$
& Base & \methodname{} & $\Delta$
& Base & \methodname{} & $\Delta$ \\
\midrule
Total Walk Distance
& 228 & 6.36 & \textbf{19.08} & $+12.72$
& 7.28 & \textbf{19.08} & $+11.80$ \\
Total Fly Distance
& 228 & 9.56 & \textbf{27.28} & $+17.72$
& 10.04 & \textbf{24.82} & $+14.78$ \\
Global How Far
& 228 & 27.19 & \textbf{39.69} & $+12.50$
& 33.25 & \textbf{44.04} & $+10.79$ \\
Global How Far From Me
& 228 & 25.57 & \textbf{34.30} & $+8.73$
& 33.99 & \textbf{43.77} & $+9.78$ \\
Global How Long
& 228 & 48.95 & \textbf{62.06} & $+13.11$
& 63.38 & \textbf{72.15} & $+8.77$ \\
Nearest From Set
& 228 & 42.54 & \textbf{46.49} & $+3.95$
& 42.98 & \textbf{46.05} & $+3.07$ \\
How Much Longer Dimension
& 228 & 34.17 & \textbf{42.28} & $+8.11$
& 37.11 & \textbf{45.88} & $+8.77$ \\
Which Longer Dimension
& 228 & 67.11 & \textbf{76.32} & $+9.21$
& 75.00 & \textbf{77.63} & $+2.63$ \\
Which Taller
& 227 & 70.04 & \textbf{70.48} & $+0.44$
& 69.16 & \textbf{70.48} & $+1.32$ \\
How Much Taller
& 226 & 30.18 & \textbf{37.48} & $+7.30$
& 35.49 & \textbf{44.25} & $+8.76$ \\
Top Higher
& 224 & 60.27 & \textbf{66.07} & $+5.80$
& 59.82 & \textbf{62.50} & $+2.68$ \\
Volume Estimation
& 168 & 35.30 & \textbf{40.12} & $+4.82$
& 14.23 & \textbf{21.85} & $+7.62$ \\
Pour Room Left
& 151 & 2.45 & \textbf{10.40} & $+7.95$
& 1.92 & \textbf{9.67} & $+7.75$ \\
Pour Leftover
& 151 & 0.86 & \textbf{6.82} & $+5.96$
& 0.86 & \textbf{6.62} & $+5.76$ \\
Gap Fit
& 62 & 72.58 & \textbf{75.81} & $+3.23$
& 56.45 & \textbf{59.68} & $+3.23$ \\
\midrule
\textbf{Overall}
& 3,033 & 35.20 & \textbf{43.70} & $+8.50$
& 37.10 & \textbf{44.50} & $+7.40$ \\
\bottomrule
\end{tabular*}
\end{table}

\subsection{Results by Subtask}
\label{app:category_results}

Tables~\ref{tab:subtask_vsi},~\ref{tab:subtask_egoprox}, and~\ref{tab:subtask_r3d} report the complete subtask results for both backbones.
Both \methodname{} students outperform their respective base models on every evaluated subtask.
The improvements span both individual spatial measurements and tasks that require combining multiple spatial observations.

For individual measurements, \methodname{}\textsubscript{\textbf{ Qwen3-VL}} and \methodname{}\textsubscript{\textbf{ SenseNova-SI}} gain 0.4 and 0.1 points on VSI-Bench's \emph{Object Size Estimation}, which asks for an object's longest dimension.
On R3D-Bench, they gain 13.1 and 8.8 points on \emph{Global How Long}, which likewise requires estimating an object's longest dimension.
These results show improvements on questions that require extracting geometric measurements from video.

We also observe substantial gains on tasks involving comparisons, aggregation, or goal-conditioned spatial relations.
On VSI-Bench, the students improve by 5.6 and 3.2 points on \emph{Object Relative Distance}, which requires comparing candidate objects by their distance to a reference.
On R3D-Bench, they gain 17.7 and 14.8 points on \emph{Total Fly Distance}, which requires combining two straight-line distance estimates.
On EgoProx, they gain 23.1 and 19.9 points on \emph{Intention Relative}, which requires interpreting head or gaze movement relative to a specified goal.
Together, these results show gains across distinct forms of spatial reasoning, including object-level estimation, relative comparison, and distance aggregation.

\subsection{Comparison with Direct-Answer Training}
\label{app:direct_answer_baseline}

We examine whether direct-answer training on the same supervision can match Dual-LoRA and the SenseNova-SI student.
All methods start from SenseNova-SI-1.3-8B.
The direct-answer baseline uses a single rank-32 LoRA on the same language-model layers and projections as Dual-LoRA, matching the total trainable adapter parameters of its two rank-16 teachers.

SFT uses the same original questions and subquestion QA from the Planner and Perception training pools, with the same answer sources and filtering criteria.
Original questions supervise only the final answer, while subquestion QA is treated as independent video question answering.
GRPO uses the same Perception samples and original multiple-choice questions as Dual-LoRA, with weak labels as reward references when subquestion ground truth is unavailable and final-answer correctness for original questions.
We use the same SFT and GRPO hyperparameters as the Dual-LoRA teachers.

We evaluate checkpoints after SFT and SFT+GRPO.
At evaluation, the baseline answers the original question directly from the video and provided answer options, without subquestions or role interaction.
For Dual-LoRA and the student, we report score differences relative to Direct-answer SFT+GRPO with 95\% confidence intervals from 10,000 paired bootstrap resamples over videos.

\begin{table}[t]
\centering
\caption{\textbf{Comparison with direct-answer training.}
For Dual-LoRA and \methodname{}, $\Delta$ reports score differences in percentage points relative to Direct-answer SFT+GRPO, with 95\% paired bootstrap confidence intervals.}
\label{tab:direct_answer_baseline}
\small
\setlength{\tabcolsep}{3pt}
\renewcommand{\arraystretch}{1.15}
\resizebox{\linewidth}{!}{%
\begin{tabular}{lcccccc}
\toprule
& \multicolumn{2}{c}{\textbf{VSI-Bench}}
& \multicolumn{2}{c}{\textbf{EgoProx}}
& \multicolumn{2}{c}{\textbf{R3D-Bench}} \\
\cmidrule(lr){2-3}
\cmidrule(lr){4-5}
\cmidrule(lr){6-7}
\textbf{Method}
& \textbf{Score} & $\Delta$ [95\% CI]
& \textbf{Score} & $\Delta$ [95\% CI]
& \textbf{Score} & $\Delta$ [95\% CI] \\
\midrule
Base
& 64.8 & --
& 34.2 & --
& 37.1 & -- \\

Direct-answer SFT
& 65.4 & --
& 39.8 & --
& 39.3 & -- \\

\rowcolor{gray!10}
Direct-answer SFT + GRPO
& 65.5 & --
& 40.7 & --
& 40.1 & -- \\

\midrule
Dual-LoRA teachers
& 67.6 & $+2.1$ [$+1.8$, $+2.6$]
& 49.0 & $+8.3$ [$+6.2$, $+10.5$]
& 44.8 & $+4.7$ [$+3.3$, $+6.1$] \\

\rowcolor{blue!8}
\methodname{}
& 66.9 & $+1.4$ [$+0.7$, $+2.2$]
& 47.2 & $+6.5$ [$+4.6$, $+8.5$]
& 44.5 & $+4.4$ [$+3.2$, $+5.6$] \\

\bottomrule
\end{tabular}%
}
\end{table}

Table~\ref{tab:direct_answer_baseline} shows that direct-answer SFT improves over the base model on all three benchmarks, and GRPO provides further gains.
Dual-LoRA nevertheless significantly outperforms Direct-answer SFT+GRPO by 2.1, 8.3, and 4.7 points on VSI-Bench, EgoProx, and R3D-Bench, respectively.
The student also exceeds this baseline by 1.4, 6.5, and 4.4 points.
Thus, the available QA supervision benefits direct answering, but does not close the gap to either trained plan-perceive-compose system.
These results support the complete pipeline over direct-answer training on the same training data and settings.

\subsection{Additional Inference Efficiency Analysis}
\label{app:efficiency}

\begin{table}[t]
\centering
\caption{\textbf{Inference cost with matched subquestion counts.}
Each group compares one student with Dual-LoRA on its own matched subset.
Metrics follow the definitions in Table~\ref{tab:inference_efficiency}.}
\label{tab:inference_efficiency_matched}
\small
\setlength{\tabcolsep}{5pt}
\renewcommand{\arraystretch}{1.1}
\begin{tabular}{lcccc}
\toprule
\textbf{Method} &
\textbf{TFLOPs} &
\shortstack{\textbf{Latency (s)}} &
\shortstack{\textbf{Generated tokens}} &
\shortstack{\textbf{Latency/token (ms)}} \\
\midrule
\rowcolor{gray!15}
\multicolumn{5}{l}{
    \textit{Subset matched to the Qwen3-VL student}
} \\
Dual-LoRA\textsubscript{\textbf{ SenseNova-SI}}
& 162.1 & 8.8 & 298.4 & 29.5 \\
\rowcolor{blue!8}
\textbf{\methodname{}}\textsubscript{\textbf{ Qwen3-VL}}
& \textbf{51.0}
& \textbf{5.8}
& \textbf{248.5}
& \textbf{23.3} \\
\midrule
\rowcolor{gray!15}
\multicolumn{5}{l}{
    \textit{Subset matched to the SenseNova-SI student}
} \\
Dual-LoRA\textsubscript{\textbf{ SenseNova-SI}}
& 156.3 & 8.6 & 277.5 & 31.0 \\
\rowcolor{blue!8}
\textbf{\methodname{}}\textsubscript{\textbf{ SenseNova-SI}}
& \textbf{49.7}
& \textbf{4.9}
& \textbf{219.0}
& \textbf{22.4} \\
\bottomrule
\end{tabular}
\end{table}

For each student, we select the evaluation questions on which it and Dual-LoRA naturally generate the same number of subquestions. 
The Qwen3-VL and SenseNova-SI subsets contain 1,110 and 2,118 questions, respectively.
Each subquestion corresponds to one Perception call.
The Qwen3-VL and SenseNova-SI subsets average 2.3 and 2.1 Perception calls per question, respectively.
Dual-LoRA costs are reported on the questions selected for each student. 

On the Qwen3-VL matched subset, Table~\ref{tab:inference_efficiency_matched} shows that the student reduces FLOPs by 68.5\% and latency from 8.8 to 5.8 seconds, a $1.52\times$ speedup. 
It generates 16.7\% fewer tokens, and end-to-end latency per generated token decreases from 29.5 to 23.3 ms. 
On the SenseNova-SI matched subset, the student reduces FLOPs by 68.2\% and latency from 8.6 to 4.9 seconds, a $1.75\times$ speedup. 
It generates 21.1\% fewer tokens, while latency per generated token decreases from 31.0 to 22.4 ms. 
Both students therefore retain an efficiency advantage when the number of subquestions is matched per original question.

\subsection{Gradient Analysis of Multi-Teacher Distillation}
\label{app:student_diagnostics}

The main results compare final task performance and student-teacher KL.
Here, we additionally examine whether the Planner (planning and composition) and Perception objectives produce conflicting update directions when learned through shared parameters.

We analyze the dense Qwen3-VL student at 12 checkpoints during MOPD.
During training, we update only the FFNs in decoder layers 24-35, corresponding to the layers converted into experts in the MoE student.
All remaining parameters are frozen.
At each checkpoint, we evaluate the same 32 pairs of PC and V samples.
For each pair, we compute the gradients of the two role losses with respect to these shared trainable parameters.
Let $g_{\mathrm{PC}}$ and $g_{\mathrm V}$ denote the resulting gradients.

Their cosine similarity is
\begin{equation}
\label{eq:role_gradient_cosine}
c =
\frac{g_{\mathrm{PC}}^{\top}g_{\mathrm V}}
{\lVert g_{\mathrm{PC}}\rVert_2
 \lVert g_{\mathrm V}\rVert_2}.
\end{equation}
A negative value indicates that the two gradients point in opposing directions: a sufficiently small gradient descent step on one role loss would increase the other to first order.

\begin{figure}[t]
\centering
\includegraphics[width=0.6\linewidth]{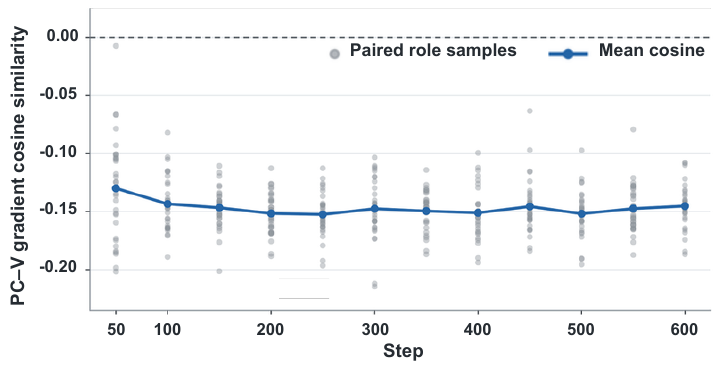}
\caption{\textbf{Gradient alignment between planning and composition (PC) and perception (V) in the dense Qwen student.}
Each gray point represents the gradient cosine similarity for one pair of PC and V samples.
Blue markers show the mean over 32 pairs at each checkpoint.
The same sample pairs are evaluated across all 12 checkpoints.}
\label{fig:role_gradient_cosine}
\end{figure}

As shown in Figure~\ref{fig:role_gradient_cosine}, all 384 measurements are negative, with an overall mean of $-0.147$.
The disagreement persists across the inspected checkpoints and sample pairs, indicating competing update directions over the evaluated parameters.
This provides additional evidence of the difficulty of learning both specialist objectives through shared parameters.

\section{Prompt Templates}
\label{app:prompts}

We provide prompts for demonstration generation, Dual-LoRA teacher inference, and student inference.
Figures~\ref{fig:planner_generation_prompt} and~\ref{fig:perception_generation_prompt} show the prompts used by DeepSeek-V4-Pro and Qwen3.5-397B.
DeepSeek generates \texttt{PLAN} demonstrations, and \texttt{DIRECT} examples are constructed as described in Appendix~\ref{app:data_sources}.
The DeepSeek \texttt{question} field includes the answer options when applicable.
The correct final answer is withheld during generation and used for verification.

The generated JSON records are converted into training conversations.
Each subquestion is formatted as an \texttt{ask\_perception} tool call, and the composition is formatted as the final Planner response.
For generated observations, the \texttt{answer} field supplies the Perception target.

Figures~\ref{fig:planner_lora_prompt} and~\ref{fig:perception_lora_prompt} show the Dual-LoRA inference prompts.
The initial tool choice implements the mode selection: \texttt{answer\_with\_base} selects \texttt{DIRECT}, while \texttt{ask\_perception} starts the \texttt{PLAN} interaction.
Subsequent calls and observations remain in the Planner's conversation history.
Each Perception call receives only the video and its current subquestion.

Figure~\ref{fig:student_prompt_template} shows the student prompt template.
Role instructions and permitted text inputs follow a shared video prefix.
Within the student, \texttt{answer\_with\_base} activates the General expert, while \texttt{ask\_perception} activates the Perception expert.
After receiving observations, the Planner expert can request further evidence or compose the final answer.
For both Dual-LoRA and student inference, the \texttt{answer\_format\_instruction} field uses the original answer format instruction provided by the corresponding benchmark.

\begingroup
\definecolor{ppblue}{RGB}{35,82,126}
\definecolor{ppgreen}{RGB}{27,112,80}
\tcbset{ppbox/.style={colback=white,colframe=black!40,
  colbacktitle=black!6,coltitle=black,fonttitle=\bfseries\small,
  boxrule=0.55pt,arc=2pt,left=8pt,right=8pt,top=5pt,bottom=5pt,
  before skip=0pt,after skip=5pt}}
\lstdefinestyle{ppjson}{basicstyle=\ttfamily\fontsize{9}{9.5}\selectfont,
  columns=fullflexible,keepspaces=true,breaklines=true,
  showstringspaces=false,aboveskip=2pt,belowskip=0pt,
  frame=none,backgroundcolor=\color{white}}

\begin{figure}[tp]
\centering
\begin{tcolorbox}[ppbox,title={System prompt}]
\fontsize{9}{10.8}\selectfont
You are a spatial reasoning planner. Determine what spatial evidence is needed to answer a question and reason from the observations provided. Do not invent observations or measurements.
\end{tcolorbox}

\begin{tcolorbox}[ppbox,title={User prompt template}]
\fontsize{9}{10.8}\selectfont
\setlength{\parindent}{0pt}\setlength{\parskip}{3pt}
\textbf{Task}

Decide the next action for the question below using the observations collected so far. Choose one action for this turn:

\hangindent=9pt\noindent-- \textcolor{ppgreen}{\texttt{ask\_perception}}: Request one or more spatial observations needed to answer the question.

\hangindent=9pt\noindent-- \textcolor{ppblue}{\texttt{compose}}: Combine the available observations into the final answer when sufficient evidence is available.

\textbf{Requirements}

\hangindent=9pt\noindent-- Request the spatial facts needed to answer the question.
Do not pass the original option list to Perception or ask it to select an option label.

\hangindent=9pt\noindent-- You may group independent subquestions in one request. If a subquestion depends on an earlier answer, wait for that answer and include the necessary information in the new subquestion.

\hangindent=9pt\noindent-- Use only the question, its options, and observations already returned. Do not treat answer options as established facts. After requesting observations, wait for their responses.

\hangindent=9pt\noindent-- For \texttt{compose}, compare measurements or perform calculations as needed, without introducing unsupported visual facts. Follow the required answer format.

\textbf{Output format}

Briefly explain your action in \texttt{thought}. Return exactly one JSON object in the corresponding format below, with no Markdown or additional text.

\noindent\begin{minipage}[t]{0.50\linewidth}
\textbf{Perception request}
\begin{lstlisting}[style=ppjson]
{
  "thought": "<brief reason>",
  "action": "ask_perception",
  "sub_questions": [
    {
      "sub_question": "<atomic question>"
    }
  ],
  "compose_answer": null
}
\end{lstlisting}
\end{minipage}\hfill
\begin{minipage}[t]{0.47\linewidth}
\textbf{Final composition}
\begin{lstlisting}[style=ppjson]
{
  "thought": "<brief reasoning>",
  "action": "compose",
  "sub_questions": [],
  "compose_answer": "<final answer>"
}
\end{lstlisting}
\end{minipage}

\textbf{Input}
\begin{lstlisting}[style=ppjson]
Question: {question}
Required answer format: {answer_format}
Interaction history: {history}
Current round: {t}
\end{lstlisting}
\end{tcolorbox}
\caption{\textbf{Prompt template for generating \texttt{PLAN} demonstrations with DeepSeek-V4-Pro.} The Planner requests spatial observations or composes an answer from the returned evidence.}
\label{fig:planner_generation_prompt}
\label{fig:planner_generation_schema}
\end{figure}
\endgroup

\begingroup
\tcbset{opbox/.style={colback=white,colframe=black!40,
  colbacktitle=black!6,coltitle=black,fonttitle=\bfseries\small,
  boxrule=0.55pt,arc=2pt,left=8pt,right=8pt,top=5pt,bottom=5pt,
  before skip=0pt,after skip=5pt}}
\lstdefinestyle{opjson}{basicstyle=\ttfamily\fontsize{9}{10.8}\selectfont,
  columns=fullflexible,keepspaces=true,breaklines=true,
  showstringspaces=false,aboveskip=2pt,belowskip=0pt,
  frame=none,backgroundcolor=\color{white}}

\begin{figure}[htp]
\centering
\begin{tcolorbox}[opbox,title={System prompt}]
\fontsize{9}{10.8}\selectfont
You answer spatial questions from video frames. Base your answer on the visual evidence and do not invent observations or measurements.
\end{tcolorbox}

\begin{tcolorbox}[opbox,title={User prompt template}]
\fontsize{9}{10.8}\selectfont
\setlength{\parindent}{0pt}\setlength{\parskip}{4pt}
\textbf{Task}

The attached frames are sampled in chronological order from a video. Answer the question below as directly as possible.

\textbf{Requirements}

\hangindent=9pt\noindent-- Answer only the requested fact, relation, or measurement, using evidence from the frames.

\hangindent=9pt\noindent-- Follow the temporal scope, reference frame, and units specified in the question.

\hangindent=9pt\noindent-- For questions about distance or size, provide an estimate when the visual evidence supports one.

\textbf{Output format}

Return exactly one JSON object with the following fields. Do not include Markdown or text outside the JSON. Use a number between 0 and 1 for \texttt{confidence}, and a JSON boolean for \texttt{answerable}.

\begin{lstlisting}[style=opjson]
{
  "answer": "<short direct answer or statement of uncertainty>",
  "confidence": <number between 0 and 1>,
  "visible_evidence": "<brief visual support or limitation>",
  "answerable": <true or false>
}
\end{lstlisting}

\textbf{Input}
\begin{lstlisting}[style=opjson]
Video frames: {video_frames}
Question: {sub_question}
\end{lstlisting}
\end{tcolorbox}
\caption{\textbf{Prompt template for Perception answer generation with Qwen3.5-397B.} The model receives video frames and one self-contained spatial subquestion, without the original answer options or planning history.}
\label{fig:perception_generation_prompt}
\end{figure}
\endgroup

\begingroup
\definecolor{lpblue}{RGB}{35,82,126}
\definecolor{lpgreen}{RGB}{27,112,80}
\tcbset{lpbox/.style={colback=white,colframe=black!40,
  colbacktitle=black!6,coltitle=black,fonttitle=\bfseries\small,
  boxrule=0.55pt,arc=2pt,left=8pt,right=8pt,top=5pt,bottom=5pt,
  before skip=0pt,after skip=5pt}}
\lstdefinestyle{lpcode}{basicstyle=\ttfamily\fontsize{9}{9.5}\selectfont,
  columns=fullflexible,keepspaces=true,breaklines=true,
  showstringspaces=false,aboveskip=2pt,belowskip=0pt,
  frame=none,backgroundcolor=\color{white}}

\begin{figure}[tp]
\centering
\begin{tcolorbox}[lpbox,title={System prompt}]
\fontsize{9}{10.8}\selectfont
You are a spatial reasoning planner for egocentric video question answering.
Determine what evidence is needed and use it to answer the question.
\end{tcolorbox}

\begin{tcolorbox}[lpbox,title={User prompt template}]
\fontsize{9}{10.8}\selectfont
\setlength{\parindent}{0pt}\setlength{\parskip}{3pt}
\textbf{Task}

Answer the question below using the video. At the beginning, choose how to answer:

\hangindent=9pt\noindent-- If the question can be answered directly from the clip as a whole, call \textcolor{lpblue}{\texttt{answer\_with\_base}}.

\hangindent=9pt\noindent-- Otherwise, call \textcolor{lpgreen}{\texttt{ask\_perception}} to obtain the spatial observations needed to answer the question.

\textbf{Requirements}

\hangindent=9pt\noindent-- You may issue multiple independent calls in one turn.
If a subquestion depends on an earlier observation, wait for that observation and include the relevant information explicitly in the new subquestion.

\hangindent=9pt\noindent-- Once you have requested observations, continue gathering evidence or compose the final answer.
Do not switch to \texttt{answer\_with\_base}.

\hangindent=9pt\noindent-- Use only the observations needed to answer the question.
When sufficient evidence is available, combine the observations and give the final answer directly in your assistant response, following the question's answer format.

\textbf{Output format}

Use the following formats for tool calls. Composition is a final assistant response.

\noindent\begin{minipage}[t]{0.50\linewidth}
\textbf{Request one observation}
\begin{lstlisting}[style=lpcode]
<tool_call>
{"name": "ask_perception",
 "arguments": {
   "sub_question": "<question>"}}
</tool_call>
\end{lstlisting}
\end{minipage}\hfill
\begin{minipage}[t]{0.47\linewidth}
\textbf{Request a direct answer}
\begin{lstlisting}[style=lpcode]
<tool_call>
{"name": "answer_with_base",
 "arguments": {}}
</tool_call>
\end{lstlisting}
\end{minipage}

\textbf{Final assistant response}
\begin{lstlisting}[style=lpcode]
<think>{composition}</think>
{final_answer}
\end{lstlisting}

\textbf{Input}

\begin{lstlisting}[style=lpcode]
<video>
Question: {question}
{options_if_applicable}
{answer_format_instruction}
\end{lstlisting}
\end{tcolorbox}
\caption{\textbf{Planner prompt for Dual-LoRA inference.}
The Planner initially chooses direct answering or perception requests.
After gathering observations, it composes the answer in its final assistant response.}
\label{fig:planner_lora_prompt}
\end{figure}
\endgroup

\begingroup
\tcbset{lvbox/.style={colback=white,colframe=black!40,
  colbacktitle=black!6,coltitle=black,fonttitle=\bfseries\small,
  boxrule=0.55pt,arc=2pt,left=8pt,right=8pt,top=5pt,bottom=5pt,
  before skip=0pt,after skip=5pt}}
\lstdefinestyle{lvcode}{basicstyle=\ttfamily\fontsize{9}{9.5}\selectfont,
  columns=fullflexible,keepspaces=true,breaklines=true,
  showstringspaces=false,aboveskip=2pt,belowskip=0pt,
  frame=none,backgroundcolor=\color{white}}

\begin{figure}[tp]
\centering
\begin{tcolorbox}[lvbox,title={System prompt}]
\fontsize{9}{10.8}\selectfont
You answer spatial questions from egocentric video.
\end{tcolorbox}

\begin{tcolorbox}[lvbox,title={User prompt template}]
\fontsize{9}{10.8}\selectfont
\setlength{\parindent}{0pt}\setlength{\parskip}{3pt}
\textbf{Task}

Answer the subquestion below using the visual evidence in the video.

\textbf{Requirements}

\hangindent=9pt\noindent-- Answer the requested fact, relation, or measurement.

\hangindent=9pt\noindent-- Follow the reference frame, temporal scope, and units specified in the subquestion.

\hangindent=9pt\noindent-- If the evidence is insufficient, state what cannot be determined.

\textbf{Output format}

Return a concise answer in plain text, following any answer format instruction below.

\textbf{Input}

\begin{lstlisting}[style=lvcode]
<video>
{sub_question}
{answer_format_instruction}
\end{lstlisting}
\end{tcolorbox}

\caption{\textbf{Perception prompt for Dual-LoRA inference.}
Each call contains the video, one self-contained subquestion, and an optional answer format instruction.
Perception receives no original answer options or planning history.}
\label{fig:perception_lora_prompt}
\end{figure}
\endgroup

\begingroup
\definecolor{spblue}{RGB}{35,82,126}
\definecolor{spgreen}{RGB}{27,112,80}
\definecolor{spgray}{RGB}{82,87,94}
\tcbset{spbox/.style={colback=white,colframe=black!35,
  colbacktitle=black!6,coltitle=black,fonttitle=\bfseries\small,
  boxrule=0.55pt,arc=2pt,left=7pt,right=7pt,top=4pt,bottom=4pt,
  before skip=0pt,after skip=5pt,
  fontupper=\fontsize{9}{10.5}\selectfont\raggedright}}
\lstdefinestyle{spcode}{basicstyle=\ttfamily\fontsize{9}{9.2}\selectfont,
  columns=fullflexible,keepspaces=true,breaklines=true,
  showstringspaces=false,aboveskip=2pt,belowskip=0pt,
  frame=none,backgroundcolor=\color{white}}

\begin{figure}[tp]
\centering
\setlength{\parindent}{0pt}
\setlength{\parskip}{3pt}

\begin{tcolorbox}[spbox,title={Shared prefix: common system instruction and video}]
You answer spatial questions about egocentric videos.
Use the available visual evidence and follow the requested answer format.
\begin{lstlisting}[style=spcode]
<video>
\end{lstlisting}
\end{tcolorbox}

\vspace{2pt}

\begin{tcolorbox}[spbox,colframe=spblue,colbacktitle=spblue!8,
  title={Planner (PC): choose a route, request evidence, and compose}]
\textbf{Role instruction}

\textit{First turn.} Decide whether the original question needs intermediate observations.
At the first turn, call \texttt{answer\_with\_base} for direct answering,
or \texttt{ask\_perception} to request evidence.

\smallskip
\textit{Subquestions.} Ask one self-contained atomic question per call.
Name the objects and specify the reference frame, temporal scope, and units when needed.
Do not pass the original option list or ask for an option label.
You may issue independent calls together.
For dependent questions, wait for the earlier answer and include the necessary information explicitly.
After issuing calls, wait for their responses.

\smallskip
\textit{After observations.} Request further evidence or combine the observations and give the final answer.
Do not switch to \texttt{answer\_with\_base} after requesting evidence.
Avoid repeating answered questions unless their answers are unclear or inconsistent.
When sufficient evidence is available, output the final answer in the requested format.

\textbf{Initial input}
\begin{lstlisting}[style=spcode]
Question: {question}
{options_if_applicable}
{answer_format_instruction}
\end{lstlisting}

\textbf{Subsequent input}
\begin{lstlisting}[style=spcode]
Observations: {subquestion_answer_pairs_in_request_order}
\end{lstlisting}
{\fontsize{8}{9.5}\selectfont
Earlier requests and observations remain in the Planner conversation.\par}

\vspace{2pt}
\noindent\begin{minipage}[t]{0.50\linewidth}
\textbf{\textcolor{spgreen}{PLAN: request an observation}}
\begin{lstlisting}[style=spcode]
<tool_call>
{"name": "ask_perception",
 "arguments": {
   "sub_question": "{sub_question}"}}
</tool_call>
\end{lstlisting}
\end{minipage}\hfill
\begin{minipage}[t]{0.47\linewidth}
\textbf{\textcolor{spgray}{DIRECT: initial turn only}}
\begin{lstlisting}[style=spcode]
<tool_call>
{"name": "answer_with_base",
 "arguments": {}}
</tool_call>
\end{lstlisting}
\end{minipage}

\vspace{2pt}
\textbf{PLAN: final composition}
\begin{lstlisting}[style=spcode]
{composition}
{final_answer}
\end{lstlisting}
\end{tcolorbox}

\noindent\begin{minipage}[t]{0.49\linewidth}
\vspace{0pt}
\begin{tcolorbox}[spbox,colframe=spgreen,colbacktitle=spgreen!8,equal height group=spbottom,
  title={Perception (V): answer one subquestion}]
\textbf{Role instruction}

Answer the current atomic spatial question using the video.
Provide the requested fact, relation, or measurement.
Follow the specified reference frame, temporal scope, units, and answer format.
Return a concise plain-text answer.
If evidence is insufficient, state what cannot be determined.

\textbf{Input after the shared prefix}
\begin{lstlisting}[style=spcode]
{sub_question}
{answer_format_instruction}
\end{lstlisting}

\textbf{Output}
\begin{lstlisting}[style=spcode]
{atomic_answer}
\end{lstlisting}

\end{tcolorbox}
\end{minipage}\hfill
\begin{minipage}[t]{0.49\linewidth}
\vspace{0pt}
\begin{tcolorbox}[spbox,colframe=spgray,colbacktitle=spgray!8,equal height group=spbottom,
  title={General (G): answer directly}]
\textbf{Role instruction}

Answer the original question directly using the video.
Do not generate subquestions or request observations.
Output the final answer in the requested format.

\textbf{Input after the shared prefix}
\begin{lstlisting}[style=spcode]
Question: {question}
{options_if_applicable}
{answer_format_instruction}
\end{lstlisting}

\textbf{Output}
\begin{lstlisting}[style=spcode]
{final_answer}
\end{lstlisting}

\end{tcolorbox}
\end{minipage}

\caption{\textbf{Student prompt template.}
All roles share a video prefix and receive separate text contexts.
The call syntax denotes internal role dispatch: \texttt{ask\_perception} activates V,
and \texttt{answer\_with\_base} activates G.
After observations return, PC can request further evidence or compose the final answer.
\texttt{DIRECT} and \texttt{PLAN} label the routes; they are not additional outputs in this template.}
\label{fig:student_prompt_template}
\end{figure}
\endgroup

\section{Qualitative Examples}
\label{app:qualitative_examples}

\begin{figure*}[t]
\centering
\includegraphics[width=\textwidth,height=0.95\textheight,keepaspectratio]{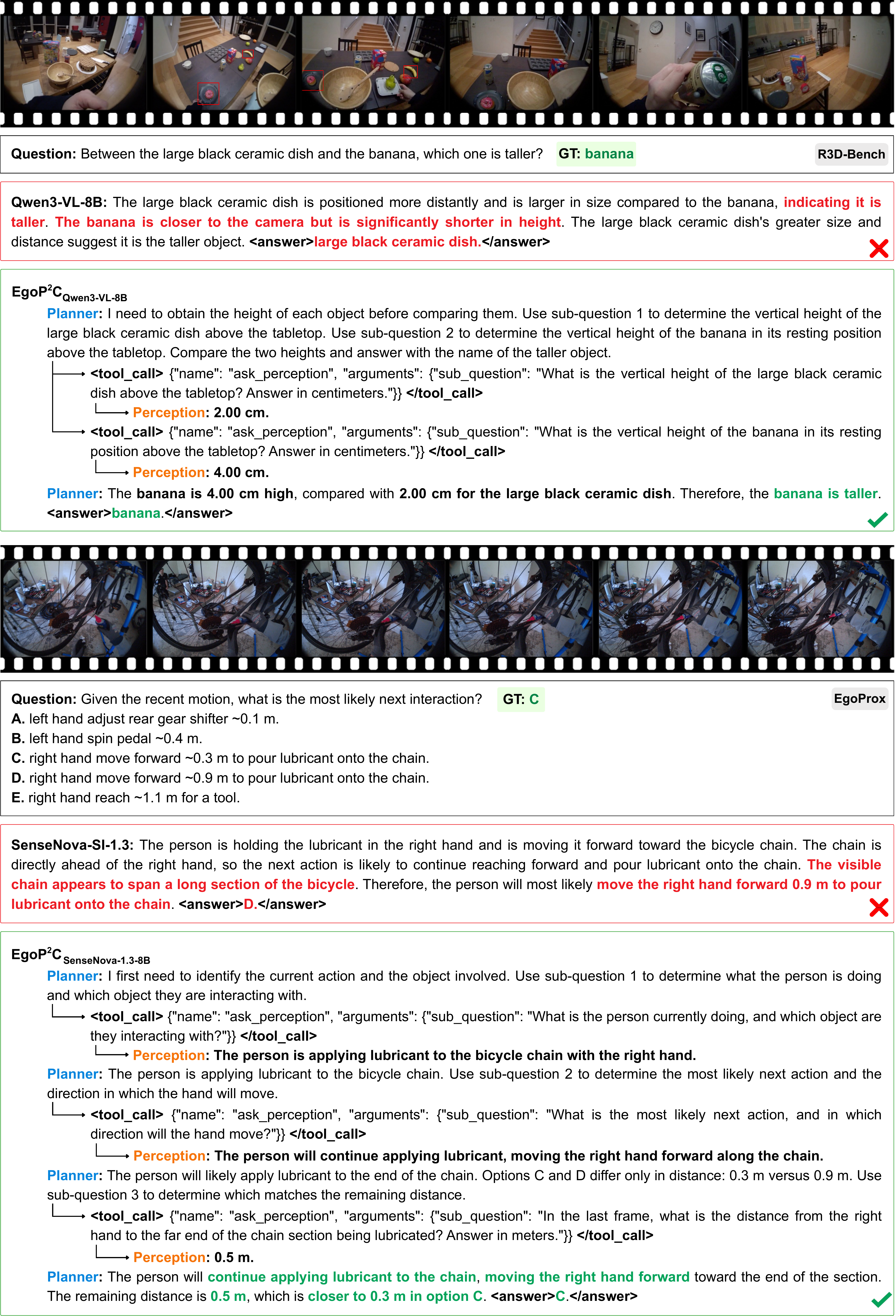}
\caption{\textbf{Illustrative comparisons on R3D-Bench (top) and EgoProx (bottom).}
\textbf{\textcolor{red}{Red text}} marks incorrect reasoning and answers, while \textbf{\textcolor{OliveGreen}{green text}} marks correct ones.}
\label{fig:examples_1}
\end{figure*}

Figure~\ref{fig:examples_1} illustrates two forms of subquestion generation.
In the R3D-Bench example (top), Qwen3-VL-8B mistakes the dish's larger overall size for greater vertical height.
\methodname{} requests the heights of both objects through \emph{independent subquestions within one planning round}.
Neither subquestion requires the other's answer, so both can be issued before receiving observations.
The returned estimates of 2\,cm for the dish and 4\,cm for the banana support the correct height comparison.

In the EgoProx example (bottom), SenseNova-SI-1.3-8B identifies the action and movement direction correctly but overestimates the remaining movement distance.
\methodname{} uses \emph{dependent subquestions across three planning rounds}.
Each subsequent question builds on observations from the preceding round.
After identifying the ongoing chain lubrication, the Planner asks about the next action and hand movement.
The predicted forward motion then motivates a question about the remaining distance to the end of the chain section in the last frame.
The Planner combines the action, direction, and distance observations to select the correct option.

\end{document}